\batchmode
\makeatletter
\def\input@path{{F:/important/doingWork/myWorks/DgS-nmf//}}
\makeatother
\documentclass[twoside,english,10pt,twocolumn,twoside]{IEEEtran}
\usepackage[T1]{fontenc}
\usepackage[utf8]{inputenc}
\usepackage{color}
\definecolor{page_backgroundcolor}{rgb}{1, 1, 1}
\pagecolor{page_backgroundcolor}
\usepackage{babel}
\usepackage{array}
\usepackage{float}
\usepackage{url}
\usepackage{multirow}
\usepackage{amsthm}
\usepackage{amsmath}
\usepackage{amssymb}
\usepackage{graphicx}
\usepackage{setspace}
\usepackage[unicode=true,
 bookmarks=true,bookmarksnumbered=true,bookmarksopen=true,bookmarksopenlevel=1,
 breaklinks=false,pdfborder={0 0 0},backref=false,colorlinks=true]
 {hyperref}
\hypersetup{pdftitle={Data guided Sparse Method for Blind Spectral Unmixing},
 pdfauthor={Fei-yun Zhu},
 citecolor=blue, urlcolor=magenta}

\makeatletter

\newcommand{\lyxmathsym}[1]{\ifmmode\begingroup\def\b@ld{bold}
  \text{\ifx\math@version\b@ld\bfseries\fi#1}\endgroup\else#1\fi}

\providecommand{\tabularnewline}{\\}
\floatstyle{ruled}
\newfloat{algorithm}{tbp}{loa}
\providecommand{\algorithmname}{Algorithm}
\floatname{algorithm}{\protect\algorithmname}

 \let\oldforeign@language\foreign@language
 \DeclareRobustCommand{\foreign@language}[1]{%
   \lowercase{\oldforeign@language{#1}}}
\theoremstyle{plain}
\newtheorem{thm}{\protect\theoremname}
\theoremstyle{definition}
\newtheorem{defn}[thm]{\protect\definitionname}
\theoremstyle{plain}
\newtheorem{lem}[thm]{\protect\lemmaname}

\ifCLASSOPTIONcompsoc
\usepackage[caption=false,font=normalsize,labelfont=sf,textfont=sf]{subfig}
\else
\usepackage[caption=false,font=footnotesize]{subfig}
\fi

\usepackage{breakurl}
\usepackage{times}
\usepackage{epsfig}
\usepackage{multirow}
\usepackage{mdwlist}
\usepackage{algorithm}
\usepackage{algorithmic}%
\usepackage{amsmath}
\usepackage{graphicx} 
\usepackage{url}

\@ifundefined{showcaptionsetup}{}{%
 \PassOptionsToPackage{caption=false}{subfig}}
\usepackage{subfig}
\makeatother

\providecommand{\definitionname}{Definition}
\providecommand{\lemmaname}{Lemma}
\providecommand{\theoremname}{Theorem}

\begin{document}
\global\long\def\mtbfA{\mathbf{A}}
 \global\long\def\mtbfa{\mathbf{a}}
 \global\long\def\mebfA{\bar{\mtbfA}}
 \global\long\def\mebfa{\bar{\mtbfa}}

\global\long\def\mhbfA{\widehat{\mathbf{A}}}
 \global\long\def\mhbfa{\widehat{\mathbf{a}}}
 \global\long\def\mtcalA{\mathcal{A}}

\global\long\def\mtbfB{\mathbf{B}}
 \global\long\def\mtbfb{\mathbf{b}}
 \global\long\def\mebfB{\bar{\mtbfB}}
 \global\long\def\mebfb{\bar{\mtbfb}}

\global\long\def\mhbfB{\widehat{\mathbf{B}}}
 \global\long\def\mhbfb{\widehat{\mathbf{b}}}
 \global\long\def\mtcalB{\mathcal{B}}

\global\long\def\mtbfC{\mathbf{C}}
 \global\long\def\mtbfc{\mathbf{c}}
 \global\long\def\mebfC{\bar{\mtbfC}}
 \global\long\def\mebfc{\bar{\mtbfc}}

\global\long\def\mhbfC{\widehat{\mathbf{C}}}
 \global\long\def\mhbfc{\widehat{\mathbf{c}}}
 \global\long\def\mtcalC{\mathcal{C}}
 \global\long\def\mtbbC{\mathbb{C}}

\global\long\def\mtbfD{\mathbf{D}}
 \global\long\def\mtbfd{\mathbf{d}}
 \global\long\def\mebfD{\bar{\mtbfD}}
 \global\long\def\mebfd{\bar{\mtbfd}}

\global\long\def\mhbfD{\widehat{\mathbf{D}}}
 \global\long\def\mhbfd{\widehat{\mathbf{d}}}
 \global\long\def\mtcalD{\mathcal{D}}

\global\long\def\mtbfE{\mathbf{E}}
 \global\long\def\mtbfe{\mathbf{e}}
 \global\long\def\mebfE{\bar{\mtbfE}}
 \global\long\def\mebfe{\bar{\mtbfe}}

\global\long\def\mhbfE{\widehat{\mathbf{E}}}
 \global\long\def\mhbfe{\widehat{\mathbf{e}}}
 \global\long\def\mtcalE{\mathcal{E}}
 \global\long\def\mtexpect{\mathbb{E}}

\global\long\def\mtbfF{\mathbf{F}}
 \global\long\def\mtbff{\mathbf{f}}
 \global\long\def\mebfF{\bar{\mathbf{F}}}
 \global\long\def\mebff{\bar{\mathbf{f}}}

\global\long\def\mhbfF{\widehat{\mathbf{F}}}
 \global\long\def\mhbff{\widehat{\mathbf{f}}}
 \global\long\def\mtcalF{\mathcal{F}}

\global\long\def\mtbfG{\mathbf{G}}
 \global\long\def\mtbfg{\mathbf{g}}
 \global\long\def\mebfG{\bar{\mathbf{G}}}
 \global\long\def\mebfg{\bar{\mathbf{g}}}

\global\long\def\mhbfG{\widehat{\mathbf{G}}}
 \global\long\def\mhbfg{\widehat{\mathbf{g}}}
 \global\long\def\mtcalG{\mathcal{G}}

\global\long\def\mtbfH{\mathbf{H}}
 \global\long\def\mtbfh{\mathbf{h}}
 \global\long\def\mebfH{\bar{\mathbf{H}}}
 \global\long\def\mebfh{\bar{\mathbf{h}}}

\global\long\def\mhbfH{\widehat{\mathbf{H}}}
 \global\long\def\mhbfh{\widehat{\mathbf{h}}}
 \global\long\def\mtcalH{\mathcal{H}}

\global\long\def\mtbfI{\mathbf{I}}
 \global\long\def\mtbfi{\mathbf{i}}
 \global\long\def\mebfI{\bar{\mathbf{I}}}
 \global\long\def\mebfi{\bar{\mathbf{i}}}

\global\long\def\mhbfI{\widehat{\mathbf{I}}}
 \global\long\def\mhbfi{\widehat{\mathbf{i}}}
 \global\long\def\mtcalI{\mathcal{I}}

\global\long\def\mtbfJ{\mathbf{J}}
 \global\long\def\mtbfj{\mathbf{j}}
 \global\long\def\mebfJ{\bar{\mathbf{J}}}
 \global\long\def\mebfj{\bar{\mathbf{j}}}

\global\long\def\mhbfJ{\widehat{\mathbf{J}}}
 \global\long\def\mhbfj{\widehat{\mathbf{j}}}
 \global\long\def\mtcalJ{\mathcal{J}}

\global\long\def\mtbfK{\mathbf{K}}
 \global\long\def\mtbfk{\mathbf{k}}
 \global\long\def\mebfK{\bar{\mathbf{K}}}
 \global\long\def\mebfk{\bar{\mathbf{k}}}

\global\long\def\mhbfK{\widehat{\mathbf{K}}}
 \global\long\def\mhbfk{\widehat{\mathbf{k}}}
 \global\long\def\mtcalK{\mathcal{K}}

\global\long\def\mtbfL{\mathbf{L}}
 \global\long\def\mtbfl{\mathbf{l}}
 \global\long\def\mebfL{\bar{\mathbf{L}}}
 \global\long\def\mebfl{\bar{\mathbf{l}}}

\global\long\def\mhbfL{\widehat{\mathbf{K}}}
 \global\long\def\mhbfl{\widehat{\mathbf{k}}}
 \global\long\def\mtcalL{\mathcal{L}}

\global\long\def\mtbfM{\mathbf{M}}
 \global\long\def\mtbfm{\mathbf{m}}
 \global\long\def\mebfM{\bar{\mathbf{M}}}
 \global\long\def\mebfm{\bar{\mathbf{m}}}

\global\long\def\mhbfM{\widehat{\mathbf{M}}}
 \global\long\def\mhbfm{\widehat{\mathbf{m}}}
 \global\long\def\mtcalM{\mathcal{M}}

\global\long\def\mtbfN{\mathbf{N}}
 \global\long\def\mtbfn{\mathbf{n}}
 \global\long\def\mebfN{\bar{\mathbf{N}}}
 \global\long\def\mebfn{\bar{\mathbf{n}}}

\global\long\def\mhbfN{\widehat{\mathbf{N}}}
 \global\long\def\mhbfn{\widehat{\mathbf{n}}}
 \global\long\def\mtcalN{\mathcal{N}}

\global\long\def\mtbfO{\mathbf{O}}
 \global\long\def\mtbfo{\mathbf{o}}
 \global\long\def\mebfO{\bar{\mathbf{O}}}
 \global\long\def\mebfo{\bar{\mathbf{o}}}

\global\long\def\mhbfO{\widehat{\mathbf{O}}}
 \global\long\def\mhbfo{\widehat{\mathbf{o}}}
 \global\long\def\mtcalO{\mathcal{O}}

\global\long\def\mtbfP{\mathbf{P}}
 \global\long\def\mtbfp{\mathbf{p}}
 \global\long\def\mebfP{\bar{\mathbf{P}}}
 \global\long\def\mebfp{\bar{\mathbf{p}}}

\global\long\def\mhbfP{\widehat{\mathbf{P}}}
 \global\long\def\mhbfp{\widehat{\mathbf{p}}}
 \global\long\def\mtcalP{\mathcal{P}}

\global\long\def\mtbfQ{\mathbf{Q}}
 \global\long\def\mtbfq{\mathbf{q}}
 \global\long\def\mebfQ{\bar{\mathbf{Q}}}
 \global\long\def\mebfq{\bar{\mathbf{q}}}

\global\long\def\mhbfQ{\widehat{\mathbf{Q}}}
 \global\long\def\mhbfq{\widehat{\mathbf{q}}}
\global\long\def\mtcalQ{\mathcal{Q}}

\global\long\def\mtbfR{\mathbf{R}}
 \global\long\def\mtbfr{\mathbf{r}}
 \global\long\def\mebfR{\bar{\mathbf{R}}}
 \global\long\def\mebfr{\bar{\mathbf{r}}}

\global\long\def\mhbfR{\widehat{\mathbf{R}}}
 \global\long\def\mhbfr{\widehat{\mathbf{r}}}
\global\long\def\mtcalR{\mathcal{R}}
 \global\long\def\mtbbR{\mathbb{R}}

\global\long\def\mtbfS{\mathbf{S}}
 \global\long\def\mtbfs{\mathbf{s}}
 \global\long\def\mebfS{\bar{\mathbf{S}}}
 \global\long\def\mebfs{\bar{\mathbf{s}}}

\global\long\def\mhbfS{\widehat{\mathbf{S}}}
 \global\long\def\mhbfs{\widehat{\mathbf{s}}}
\global\long\def\mtcalS{\mathcal{S}}

\global\long\def\mtbfT{\mathbf{T}}
 \global\long\def\mtbft{\mathbf{t}}
 \global\long\def\mebfT{\bar{\mathbf{T}}}
 \global\long\def\mebft{\bar{\mathbf{t}}}

\global\long\def\mhbfT{\widehat{\mathbf{T}}}
 \global\long\def\mhbft{\widehat{\mathbf{t}}}
 \global\long\def\mtcalT{\mathcal{T}}

\global\long\def\mtbfU{\mathbf{U}}
 \global\long\def\mtbfu{\mathbf{u}}
 \global\long\def\mebfU{\bar{\mathbf{U}}}
 \global\long\def\mebfu{\bar{\mathbf{u}}}

\global\long\def\mhbfU{\widehat{\mathbf{U}}}
 \global\long\def\mhbfu{\widehat{\mathbf{u}}}
 \global\long\def\mtcalU{\mathcal{U}}

\global\long\def\mtbfV{\mathbf{V}}
 \global\long\def\mtbfv{\mathbf{v}}
 \global\long\def\mebfV{\bar{\mathbf{V}}}
 \global\long\def\mebfv{\bar{\mathbf{v}}}

\global\long\def\mhbfV{\widehat{\mathbf{V}}}
 \global\long\def\mhbfv{\widehat{\mathbf{v}}}
\global\long\def\mtcalV{\mathcal{V}}

\global\long\def\mtbfW{\mathbf{W}}
 \global\long\def\mtbfw{\mathbf{w}}
 \global\long\def\mebfW{\bar{\mathbf{W}}}
 \global\long\def\mebfw{\bar{\mathbf{w}}}

\global\long\def\mhbfW{\widehat{\mathbf{W}}}
 \global\long\def\mhbfw{\widehat{\mathbf{w}}}
 \global\long\def\mtcalW{\mathcal{W}}

\global\long\def\mtbfX{\mathbf{X}}
 \global\long\def\mtbfx{\mathbf{x}}
 \global\long\def\mebfX{\bar{\mtbfX}}
 \global\long\def\mebfx{\bar{\mtbfx}}

\global\long\def\mhbfX{\widehat{\mathbf{X}}}
 \global\long\def\mhbfx{\widehat{\mathbf{x}}}
 \global\long\def\mtcalX{\mathcal{X}}

\global\long\def\mtbfY{\mathbf{Y}}
 \global\long\def\mtbfy{\mathbf{y}}
\global\long\def\mebfY{\bar{\mathbf{Y}}}
 \global\long\def\mebfy{\bar{\mathbf{y}}}

\global\long\def\mhbfY{\widehat{\mathbf{Y}}}
 \global\long\def\mhbfy{\widehat{\mathbf{y}}}
 \global\long\def\mtcalY{\mathcal{Y}}

\global\long\def\mtbfZ{\mathbf{Z}}
 \global\long\def\mtbfz{\mathbf{z}}
 \global\long\def\mebfZ{\bar{\mathbf{Z}}}
 \global\long\def\mebfz{\bar{\mathbf{z}}}

\global\long\def\mhbfZ{\widehat{\mathbf{Z}}}
 \global\long\def\mhbfz{\widehat{\mathbf{z}}}
\global\long\def\mtcalZ{\mathcal{Z}}

\global\long\def\mtvar{\mathbf{\text{Var}}}

\global\long\def\mtth{\text{th}}

\global\long\def\mtbfzero{\mathbf{0}}
 \global\long\def\mtbfone{\mathbf{1}}

\global\long\def\mttrace{\text{Tr}}

\global\long\def\mttotalVariation{\text{TV}}

\global\long\def\mtdet{\text{Det}}

\global\long\def\mtvec{\mathbf{\text{vec}}}

\global\long\def\mtvar{\mathbf{\text{var}}}

\global\long\def\mtcov{\mathbf{\text{cov}}}

\global\long\def\mtsubTo{\mathbf{\text{s.t.}}}

\global\long\def\mtfor{\text{for}}

\global\long\def\mtrank{\text{rank}}

\global\long\def\mtdiag{\mathbf{\text{diag}}}

\global\long\def\mtsign{\mathbf{\text{sign}}}

\global\long\def\mtloss{\mathbf{\text{loss}}}

\global\long\def\mtwhen{\text{when}}

\global\long\def\mtexpect{\mathbb{E}}

\global\long\def\mtcalN{\mathcal{N}}

\global\long\def\mtbbR{\mathbb{R}}

\title{Spectral Unmixing via Data-guided Sparsity}

\author{Feiyun~Zhu, Ying~Wang, Bin~Fan, Shiming~Xiang, Gaofeng\ Meng
and~Chunhong~Pan%
\thanks{Feiyun~Zhu, Ying~Wang, Bin~Fan, Shiming~Xiang, Gaofeng\ Meng
and~Chunhong~Pan are with the National Laboratory of Pattern Recognition,
Institute of Automation, Chinese Academy of Sciences (e-mail: \{fyzhu,
ywang, bfan, smxiang, gfmeng and chpan\}@nlpr.ia.ac.cn).%
}}

\markboth{IEEE Transactions on Image Processing}{F. Y. Zhu \MakeLowercase{\emph{et al.}}:
DgS-NMF for Spectral Unmixing}
\maketitle
\begin{abstract}
Hyperspectral unmixing, the process of estimating a common set of
spectral bases and their corresponding composite percentages at each
pixel, is an important task for hyperspectral analysis, visualization
and understanding. From an unsupervised learning perspective, this
problem is very challenging---both the spectral bases and their composite
percentages are unknown, making the solution space too large. To reduce
the solution space, many approaches have been proposed by exploiting
various priors. In practice, these priors would easily lead to some
unsuitable solution. This is because they are achieved by applying
an identical strength of constraints to all the factors, which does
not hold in practice. To overcome this limitation, we propose a novel
sparsity based method by learning a data-guided map to describe the
individual mixed level of each pixel. Through this data-guided map,
the $\ell_{p}\left(0<p<1\right)$ constraint is applied in an adaptive
manner. Such implementation not only meets the practical situation,
but also guides the spectral bases toward the pixels under highly
sparse constraint. What's more, an elegant optimization scheme as
well as its convergence proof have been provided in this paper. Extensive
experiments on several datasets also demonstrate that the data-guided
map is feasible, and high quality unmixing results could be obtained
by our method.\end{abstract}

\section{Introduction}

\begin{figure}[t]
\centering{}\includegraphics[width=0.96\columnwidth]{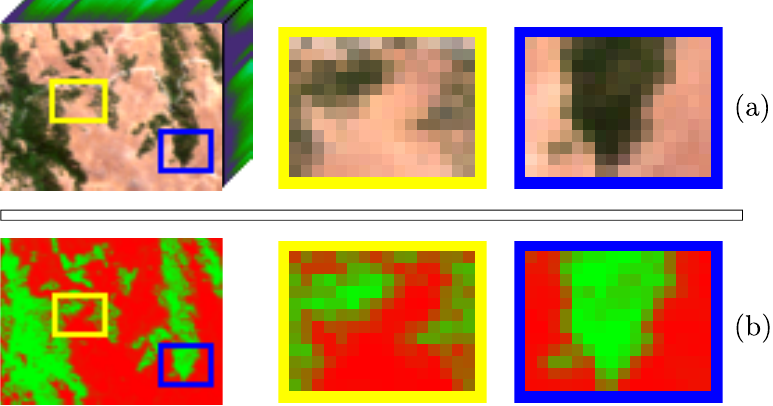}\caption{Two observations behind the figure: the mixed level of each pixel
varies over image grids; the pixels in the transition area are more
likely to be highly mixed. (a) Hyperspectral image and its close-ups.
(b) \emph{Abundances} of two substances in (a), indicated by the proportions
of red and green inks.\label{fig:illustrate_individual_mixedLevel}}
\end{figure}

\IEEEPARstart{H}{yperspectral} imaging, the process of capturing
a 3D image cube at hundreds of contiguous and narrow spectral channels,
has been used in a wide range of fields~\cite{Keshava_03_LJ_unmixingSurvey,Dias_12_AEORS_HUoverview}.
Although this type of images contains substantial information, there
are two underlying ``problems''. One ``problem'' is that as a 3D image
cube, it is very hard for computers to display\ \cite{ShangshuCai_2007_TGRS_hyperVisualizationUsingDoubleLayers},
thus hampering human to understand this type of images. Another ``problem''
is called ``mixed'' pixels---due to the low spatial resolution of
hyperspectral sensors, the spectra of different substances would unavoidably
blend together~\cite{Keshava_03_LJ_unmixingSurvey,fyzhu_2014_IJPRS_SS_NMF,Keshava_2002_SignalProcessingMagazine_SpectralUnmixing},
yielding a great number of mixed pixels as shown in Fig.\,\ref{fig:illustrate_individual_mixedLevel}.
 To address the above two problems,  various Hyperspectral Unmixing
(HU) methods have been proposed. What is more, HU is essential for
various hyperspectral applications, such as sub-pixel mapping~\cite{Mertens_2003_IJRS_subPixelMapping},
hyperspectral enhancement~\cite{zhGuo_2009_SPIE_L1unmixing&Enhancement},
high-resolution hyperspectral imaging~\cite{Kawakami_2011_CVPR_highResolutionHyperImaging},
detection and identification of ground targets~\cite{Qian_11_TGRS_NMF+l1/2}.

Formally, the HU method takes in a hyperspectral image with $L$ channels
and assumes that each pixel spectrum $\mtbfy$ is a composite of $K$
spectral bases $\left\{ \mtbfm_{k}\right\} _{k=1}^{K}\!\in\!\mtbbR_{+}^{L}$~\cite{Dias_12_AEORS_HUoverview,cbLi_2012_Tip_CompressSensing&Unmixing,XLu_2013_TGRS_ManifoldSparseNMF}.
Each spectral base is called an \emph{endmember}, representing the
pure spectrum, such as the spectra of ``water'', ``grass'' etc. Specifically,
the pixel spectrum $\mtbfy$ is generally approximated by a nonnegative
linear combination as 
\begin{equation}
\mtbfy=\sum_{k=1}^{K}\mtbfm_{k}a_{k},\quad\mtsubTo\: a_{k}\geq0\ \text{and\ }\sum_{k=1}^{K}a_{k}=1,\label{eq:linearMixtureModel}
\end{equation}
where $a_{k}$ is the composite percentage (i.e. \emph{abundance})
of the $k^{\mtth}$ \emph{endmember}. In the unsupervised setting,
both \emph{endmembers} $\left\{ \mtbfm_{k}\right\} _{k=1}^{K}$ and
\emph{abundances} $\left\{ a_{k}\right\} _{k=1}^{K}$ are unknown.
Such case makes the solution space really large~\cite{LiuXueSong_2011_TGRS_ConstrainedNMF}.
Prior knowledge is required to restrict the solution space, or even
to bias the solution toward good results. 

To shrink the solution space, many methods have been proposed by exploiting
various constraints on \emph{abundances}\,\cite{Qian_11_TGRS_NMF+l1/2,XLu_2013_TGRS_ManifoldSparseNMF}
and \emph{endmembers}\,\cite{Miao_07_ITGRS_NMFMVC,nWang_13_SelectedTopics_EDC-NMF}.
Specifically, the sparse constraints~\cite{Qian_11_TGRS_NMF+l1/2,LiuXueSong_2011_TGRS_ConstrainedNMF}
and the spatial constraints~\cite{XLu_2013_TGRS_ManifoldSparseNMF,JmLiu_12_SlectedTopics_W-NMF}
are the most popular ones. Unfortunately, all these methods exploit
an identical strength of constraints on all the factors, which may
not meet the practical situation. An example is illustrated in Fig.\,\ref{fig:illustrate_individual_mixedLevel},
where the mixed level%
\footnote{Note that a pixel with higher mixed levels should own the \emph{abundance
}vectors of lower sparse levels, and vice versa.\label{fn:mixedPixel_sparseAbundance}%
} of each pixel varies over image grids. Such an example indicates
that it is better to impose the sparse constraint of adaptive strengths
for the pixels.

In this paper, we propose a Data-guided Sparsity regularized Nonnegative
Matrix Factorization (DgS-NMF) method for the HU task. The basic motivation
is that the mixed level of each pixel might be different from each
other, as shown in Fig.~\ref{fig:illustrate_individual_mixedLevel}.
To give a more accurate model, a  data-guided map (DgMap) is incorporated
into the NMF framework so as to adaptively impose the sparse constraint
for each pixel. First, via a two step strategy, the DgMap is learned
from the hyperspectral image, describing the mixed level of each pixel.
Given this DgMap, the $\ell_{p}\left(0<p<1\right)$-norm based sparsity
constraint is individually imposed. For each pixel, the choice of
$p$ is totally dependent on the corresponding DgMap value. Such case
is better suited to the practical situation, thus expected to achieve
better HU results. Besides, this adaptive sparsity constraint would
influence the estimation of \emph{endmembers}, potentially, guiding
the \emph{endmembers} toward the pixels under highly sparse constraints.
Extensive empirical results verify that our method is highly promising
for the HU task.

The rest of this paper is organized as follows: in Section~\ref{sec:PreviousWork},
we briefly review several recent HU methods. Section~\ref{sec:Data-guided-Map}
presents how to learn DgMaps from the original hyperspectral image
cube. The DgS-NMF method as well as its properties are given in Section~\ref{sec:Data-guided-Sparse-NMF}.
Then, extensive experiments and detailed comparisons are provided
in Section~\ref{sec:Evaluation}. Finally, the conclusion of this
work is drawn in Section~\ref{sec:Conclusions}.

\section{\label{sec:PreviousWork}Previous Work}

The HU methods could be typically categorized into two types: geometric
methods~\cite{Jose_05_TGRS_Vca,Chang_06_ITGRS_EndmemberExt,Martin_12_SelectedTopics_SSPP}
and statistical ones~\cite{cbLi_2012_Tip_CompressSensing&Unmixing,Dobigeon_ISP_BaysianHU,Jose_09_WHISPERS_SplittingAugmentedLag,Jose_12_TGRS_HUbyDirchletComponents}.
Usually, the geometric methods utilize a simplex to describe the distribution
of hyperspectral pixels. The vertices of this simplex are viewed as
the \emph{endmembers}. Perhaps, N-FINDR~\cite{Michael_99_PSCIS_nFindr}
and Vertex Component Analysis (VCA)~\cite{Jose_05_TGRS_Vca} are
the most popular geometric methods. In N-FINDR, the \emph{endmembers}
are identified by inflating a simplex inside the hyperspectral pixel
distribution and treating the vertices of a simplex with the largest
volume as \emph{endmembers}~\cite{Michael_99_PSCIS_nFindr}. While
VCA~\cite{Jose_05_TGRS_Vca} projects all pixels onto a direction
orthogonal to the simplex spanned by the chosen\emph{ endmembers};
the new \emph{endmember} is identified as the extreme of the projection.
Although these methods are simple and fast, they suffer from the requirement
of pure pixels for each \emph{endmember}, which is usually unavailable
in practice. 

Accordingly, a number of statistical methods have been proposed for
or applied to the HU task, among which the Nonnegative Matrix Factorization
(NMF)~\cite{Lee_99_Nature_NMF} and its variants are the most popular
ones. As an unsupervised method, the goal of NMF is to find two nonnegative
matrices to approximate the original matrix with their product~\cite{Cai_11_PAMI_GNMF}.
Specifically, the nonnegative constraint on the two factor matrices
only allows additive combinations, not subtractions, resulting in
a parts-based representation. This parts-based property could ensure
the representation results to be more intuitive and interpretable,
since psychological and physiological evidences have shown that human
brain works in a parts-based way~\cite{Palmer_77_CP_perceptualRepresent,Logothetis_96_ARN_VisObjRecognition}. 

Although the NMF method is well suited to many applications, such
as face analysis~\cite{Stanzli_01_CVPR_locNMF,Roman_11_PAMI_EarthMoverNMF}
and documents clustering~\cite{WeiXu_03_SIGIR_DocClusterNMF,Farial_06_InfProManag_DocumNMF},
the objective function of NMF is non-convex, inherently resulting
in large solution space~\cite{Lee_00_NIPS_NMF}. Many extensions
have been proposed by exploiting various priors to restrict the solution
space. For the HU problem, these priors are either imposed to the
\emph{abundance} matrix or to the\emph{ endmember} matrix. For example,
the Local Neighborhood Weights regularized NMF method (W-NMF)~\cite{JmLiu_12_SlectedTopics_W-NMF}
assumes that the hyperspectral pixels are on a manifold structure,
which could be transferred to the \emph{abundance} space through a
Laplace graph constraint. Actually, this constraint has a smooth influence,
and eventually weaken the parts-based property of NMF. 

Inspired by the MVC-NMF~\cite{Miao_07_ITGRS_NMFMVC} method, Wan
et al.~\cite{nWang_13_SelectedTopics_EDC-NMF} proposed the EDC-NMF
method. The basic assumption is that due to the high spectral resolution
of sensors, the \emph{endmember} spectra should be smooth itself and
different as much as possible from each other. However, in their algorithm,
they take a derivative of \emph{endmembers}, introducing negative
values to the updating rules. To make up this drawback, the elements
in the \emph{endmember} matrix are required to project to a given
nonnegative value after each iteration. Consequently, the regularization
parameters could not be chosen freely, limiting the efficacy of this
method. 

Other algorithms assume that in hyperspectral images most pixels are
mixed by only a few \emph{endmembers}, hence exploiting various kinds
of sparse constraints on the \emph{abundance}\ \cite{fyzhu_2014_IJPRS_SS_NMF}.
Specifically, the $\ell_{1/2}$-NMF~\cite{Qian_11_TGRS_NMF+l1/2}
is a very popular sparsity regularized NMF method. It is an improvement
from Hoyer's lasso regularized NMF method~\cite{Hoyer_02_NNSP_NMF_l1}.
There are two advantages of the $\ell_{1/2}$-NMF over the lasso regularized
NMF. One advantage is that the lasso constraint~\cite{Tibshirani_94_Statist_Lasso,Donoho_96_ITIT_CS}
could not enforce further sparse when the full additivity constraint
is used, limiting the effectiveness of this method~\cite{Qian_11_TGRS_NMF+l1/2}.
Another advantage  is that Fan et al.~\cite{Fan_2001_JASA_LaNorm}
has proven that the $\ell_{p}\left(0<p<1\right)$ constraint could
obtain sparser solutions than the $\ell_{1}$ norm does.

Our method is also derived from the sparse assumption on the \emph{abundance}.
Different from the existing methods, the strength of sparse constraints
is learned from the data itself and applied in an adaptive way. Such
improvement not only meets the practical situation better, but also
help the optimization process to reach a more suitable local minimum.

\section{\label{sec:Data-guided-Map}Data-guided Map (DgMap)}

Generally, the Data-guided Map (DgMap) is a map learnt from the hyperspectral
image that describes the strength of priors (constraints) for each
factor. In this work, the DgMap depicts the mixed level of each pixel.
It comes from two observations that: 1) in the local image window,
the mixed level of each pixel might be more or less different from
each other as shown in Fig.~\ref{fig:illustrate_individual_mixedLevel};
2) in the whole image, the pixels in the transition area are very
likely to be highly mixed (c.f. Footnote\ \ref{fn:mixedPixel_sparseAbundance}).
For the second idea, Fig.~\ref{fig:illustrate_individual_mixedLevel}
illustrates an example, where there are two\emph{ }targets (i.e. ``tree''
and ``soil'') in the scene. The pixels in the transition area are
very likely to be mixed by spatially neighboring pixels from these
two targets, thus, yielding a great number of mixed pixels. Therefore,
these pixels in the transition area should receive weaker sparse constraint
than pixels in the other areas. In the following, we would elaborate
how to learn such a DgMap from the hyperspectral image via a two step
strategy.

\subsection{Initial Data-guided Map \label{sub:InitialDataGuided_Map}}

Suppose we are given a hyperspectral image $\left\{ \mtbfy_{n}\right\} _{n=1}^{N}\!\in\!\mtbbR_{+}^{L}$
with $N$ pixels and $L$ channels. It is reasonable to assume that
the pixels in the transition area are more or less different from
their spatial neighbors as shown in Fig.\,\ref{fig:illustrate_individual_mixedLevel}.
For this reason, the initial DgMap $\mtbfh^{\left(0\right)}\!\in\!\mtbbR_{+}^{N}$
could be learnt by measuring the uniformity of neighboring pixels
over the entire image, i.e. $\mtbfh^{\left(0\right)}\!=\! f\left(\mtbfy_{1},\cdots,\mtbfy_{N}\right)$.
In this way, the inhomogeneous areas are treated as the transition
ones. For the $i^{\mtth}$ pixel, its value in the DgMap could be
estimated by measuring the similarity between spatially neighboring
pixels as follows: 
\begin{equation}
h_{i}^{\left(0\right)}=\sum_{j\in\mtcalN_{i}}s_{ij},\label{eq:initialDgMap}
\end{equation}
where $\mtcalN_{i}$ is the neighborhood of the $i^{\mtth}$ pixel
that includes four neighbors; $s_{ij}$ is the similarity between
the $i^{\mtth}$ pixel and its neighboring pixel $\mtbfy_{j}$ by
the dot-product metric

\begin{equation}
s_{ij}=\frac{\mtbfy_{i}^{T}\mtbfy_{j}}{\left\Vert \mtbfy_{i}\right\Vert \cdot\left\Vert \mtbfy_{j}\right\Vert },\label{eq:dotProduct_similarity}
\end{equation}
which is a classic measure in the HU study~\cite{Qian_11_TGRS_NMF+l1/2,LiuXueSong_2011_TGRS_ConstrainedNMF},
or by the heat kernel similarity metric 
\begin{equation}
s_{ij}=\exp\left(-\frac{\left\Vert \mtbfy_{j}-\mtbfy_{i}\right\Vert _{2}^{2}}{\sigma}\right).\label{eq:heatKernel_similarity}
\end{equation}
The value of $\sigma$ controls the constrast of DgMaps. Generally,
a smaller $\sigma$ results in a DgMap with higher constrast. In all
the experiments, $\sigma$ is set as $\sigma\in\left[0.005,\,0.08\right]$.

\begin{figure}[tb]
\centering{}\includegraphics[width=0.99\columnwidth]{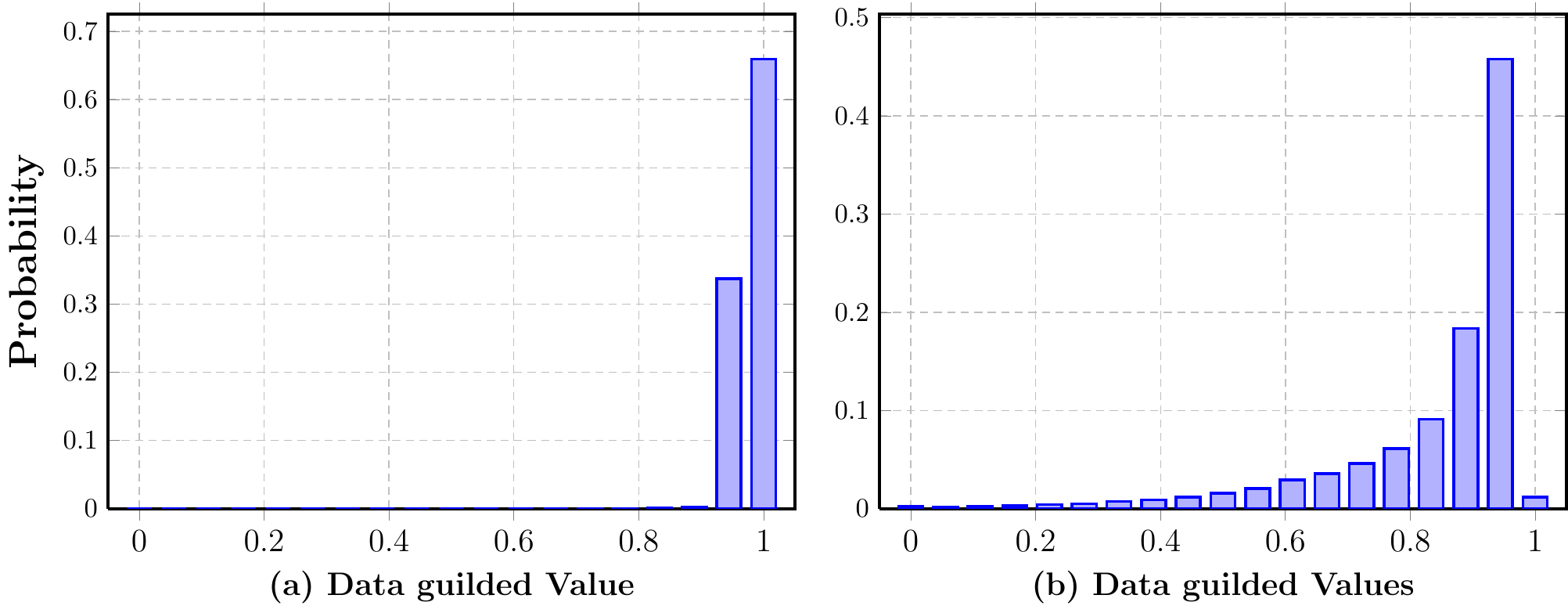}\vspace{-0.2cm}\caption{Histogram of DgMap values learnt from 36 hyperspectral images respectively
by: (a) dot-product measure~\eqref{eq:dotProduct_similarity} and
(b) heat kernel measure~\eqref{eq:heatKernel_similarity}. \label{fig:Histogram_of_twoMeasures}}
\vspace{-0.4cm}
\end{figure}

To evaluate the effectiveness of the definition of DgMaps in~\eqref{eq:initialDgMap},
we collect a set of 36 hyperspectral images%
\footnote{This image set includes hyperspectral scenes of urban areas, suburbs
areas, farmland areas, mine areas, airports and so on. In average,
there are $369\times369$ pixels and $193$  channels in a hyperspectral
image. %
} and calculate their DgMaps according to the similarity measures in~\eqref{eq:dotProduct_similarity}
and~\eqref{eq:heatKernel_similarity}. The results are plotted in
Fig.~\ref{fig:Histogram_of_twoMeasures}, showing the histogram of
DgMap values over all the 36 images. As Fig.~\ref{fig:Histogram_of_twoMeasures}a
shows, it is of little  effect by using the dot-product measure---up
to $99.61\%$ guided values are located in a narrow range of $\left[0.95,1\right]$.
Such case suggests that almost all the pixels have similar DgMap values,
lacking of guided information. Contrarily, the DgMaps from the heat
kernel measure contain more information, as shown in Figs.~\ref{fig:Histogram_of_twoMeasures}b.
Therefore, we choose the heat kernel measure to learn the initial
DgMap.

\subsection{\label{sub:FineTuned-DgMap}Fine Tuned Data-guided Map }

Through the local uniformity assumption and the heat kernel measure,
the learned DgMap does not have the global consistency over the entire
image. Therefore, we further propose a fine tuning step to refine
the initial DgMap. For this purpose, the closed-form method~\cite{ALevin_2008_PAMI_closedFormd,kaiMingHe_2011_PAMI_darkChannel}
is adopted. The advantages are in two folds: 1) this fine tuning process
not only propagates the guidance information over the entire image,
but also maintains the structures latent in the original hyperspectral
data cube\ \cite{ALevin_2006_CVPR_closedFormd,guQuanQuan_2011_AAAI_MultiTaskClustering};
2) according to our experiments, the fine tuned DgMap further improves
the HU performance although the initial DgMap already outperforms
the state-of-the-art.

Specifically, the closed-form method is based on the assumption that
in each small window, the data-guided values come from the same projection
as~\cite{ALevin_2008_PAMI_closedFormd}:
\[
h_{j}=\mtbfw_{i}^{T}\mtbfy_{j}+b_{i},\quad\text{for}\ j\in\mtcalN_{i},i\in\left\{ 1,\cdots,N\right\} ,
\]
 where $\mtbfy_{i}$ is the $i^{\mtth}$ pixel; $\mtcalN_{i}$ is
the eighborhood of $\mtbfy_{i}$%
\footnote{Note that the neighborhood $\mtcalN_{i}$ defined here is different
from the neighborhood $\mtcalN_{i}$ used in Section\ \ref{sub:InitialDataGuided_Map}.%
}; $\mtbfw_{i}$ is the projection vector and $b_{i}$ is a bias term.
The local adjustment from the initial DgMap $\left\{ h_{j}^{\left(0\right)}\right\} _{j=1}^{N}$
is formulated as
\begin{align*}
h_{j}^{*}\leftarrow & \arg\min_{h_{j}}\left(\alpha\left(h_{j}-h_{j}^{\left(0\right)}\right)^{2}+\left(h_{j}-\mtbfw_{i}^{T}\mtbfy_{j}-b_{i}\right)^{2}\right),\\
 & \forall\, j\in\mtcalN_{i},i\in\left\{ 1,2,\cdots,N\right\} .
\end{align*}
 The local window is placed in an overlapping manner. This case ensures
the property of propagating guidance information between neighboring
pixels~\cite{ALevin_2008_PAMI_closedFormd}. The bigger the local
window is, the wider the propagation could spread. Besides, in each
 local window, the gradient field of the DgMap is linearly related
to the corresponding image gradient field as $\nabla h_{j}=\mtbfw_{i}^{T}\nabla\mtbfy_{j},\forall j\!\in\!\mtcalN_{i},i\!\in\!\left\{ 1,\cdots,N\right\} $,
transferring the gradient distributions as well as the transition
information latent in the original hyperspectral image into the newly
learnt DgMap~\cite{ShanQi_2010_TVCG_ToneMapping,guQuanQuan_2011_AAAI_MultiTaskClustering}.
As a result, we could refine the DgMap according to the structures
latent in the original image cube.

Considering all the local minimizing problems together as well as
the numerical stability, we can fine tune the DgMap by minimizing
the following quadratic function~\cite{ShanQi_2010_TVCG_ToneMapping,ALevin_2008_PAMI_closedFormd}:
\begin{align}
E\left(\mtbfh,\mtbfw,b\right)= & \alpha\left\Vert \mtbfh-\mtbfh^{\left(0\right)}\right\Vert _{2}^{2}+\sum_{i=1}^{N}\Biggl(\epsilon\left\Vert \mtbfw_{i}\right\Vert _{2}^{2}\nonumber \\
 & +\sum_{j\in\mtcalN_{i}}\left(h_{j}-\mtbfw_{i}^{T}\mtbfy_{j}-b_{i}\right)^{2}\Biggr),\label{eq:globalGraphErrorWithConstraint}
\end{align}
where $\mtbfh^{\left(0\right)}\!=\!\left[h_{1}^{\left(0\right)},\cdots,h_{N}^{\left(0\right)}\right]^{T}\!\in\!\mtbbR_{+}^{N}$
is the initial DgMap; $\epsilon\in[10^{-7},10^{-4}]$ controls the
smooth level of the refined DgMap $\mtbfh$; $\alpha\in\left[10^{-6},10^{-4}\right]$
controls the strength of the fine tuning process.  A smaller $\alpha$
corresponds to a stronger refinement.

The objective function above could be further simplified by setting
$\frac{\partial E}{\partial\mtbfw}=0,\frac{\partial E}{\partial b}=0$
and substituting their solutions into~\eqref{eq:globalGraphErrorWithConstraint},
yielding a compact objective function as: 
\begin{equation}
E\left(\mtbfh\right)=\alpha\left\Vert \mtbfh-\mtbfh^{\left(0\right)}\right\Vert _{2}^{2}+\mtbfh^{T}\mtbfL\mtbfh,\label{eq:globalGraphErrorSimple}
\end{equation}
 where $\mtbfL$ is a highly sparse matrix that has been proven to
be a graph Laplacian by~\cite{Cvetkovic_1980_Book_SpectraOfGraphs}.
It is defined as 
\begin{equation}
\mtbfL=\sum_{n=1}^{N}\mtbfS_{i}^{T}\mtbfL_{i}\mtbfS_{i},\label{eq:learningAffinityMatrix}
\end{equation}
 where $\mtbfS_{i}^{T}$ is the $i^{\mtth}$  column selection matrix
that selects the pixels in the $i^{\mtth}$  local window from the
whole hyperspectral image as $\mtbfY_{i}\!=\!\mtbfY\mtbfS_{i}^{T}$,
$\mtbfL_{i}=\mtbfG_{i}\mtbfG_{i}$, in which $\mtbfG_{i}\!=\!\left(\mtbfP-\mebfY_{i}^{T}\left(\mebfY_{i}\mebfY_{i}^{T}+\epsilon\mtbfI\right)^{-1}\mebfY_{i}\right)$,
$\mtbfP\!=\!\mtbfI-\frac{1}{\left|\mtcalN_{i}\right|}\mtbfone\mtbfone^{T}$
is the centering matrix with $\left|\mtcalN_{i}\right|\!\times\!\left|\mtcalN_{i}\right|$
elements and $\mebfY_{i}\!=\!\mtbfY_{i}\mtbfP$ contains the zero
mean pixels in $\mtcalN_{i}$~\cite{smXiang_2010_TIP_TurboPixelEigenImage,smXiang_2011_TSMBC_LLE_LTSA}.
\begin{figure}[tb]
\begin{singlespace}
\noindent \begin{centering}
\subfloat[\label{fig:fineTunedDgMap-1}]{\centering{}\includegraphics[width=0.237\columnwidth]{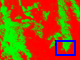}}\ \subfloat[\label{fig:fineTunedDgMap-2}]{\centering{}\includegraphics[width=0.237\columnwidth]{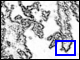}}\ \subfloat[\label{fig:fineTunedDgMap-3}]{\centering{}\includegraphics[width=0.237\columnwidth]{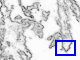}}\ \subfloat[\label{fig:fineTunedDgMap-4}]{\centering{}\includegraphics[width=0.237\columnwidth]{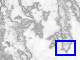}}
\par\end{centering}

\noindent \begin{centering}
\subfloat[]{\centering{}\includegraphics[width=0.237\columnwidth]{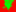}}\ \subfloat[]{\centering{}\includegraphics[width=0.237\columnwidth]{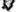}}\ \subfloat[]{\centering{}\includegraphics[width=0.237\columnwidth]{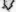}}\ \subfloat[]{\centering{}\includegraphics[width=0.237\columnwidth]{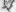}}
\par\end{centering}
\end{singlespace}

\noindent \centering{}\caption{(a) The \emph{abundance} map of the hyperspectral image in Fig.~\ref{fig:illustrate_individual_mixedLevel},
where the proportions of red and green inks represent the \emph{abundances}
of two targets. (b) The initial DgMap from the heat kernel measure.
(c) Fine tuned DgMap learnt by $3\times3$ window. (d) Fine tuned
DgMap learnt by $7\times7$ window. (e)-(h) are the close ups of (a)-(d)
respectively. (Best viewed in color) \label{fig:fineTunedDgMap}}
\end{figure}

Since the objective function~\eqref{eq:globalGraphErrorSimple} is
quadratic in $\mtbfh$, it can be solved by setting the derivative
to zero as $\nabla E\left(\mtbfh\right)=\mtbfzero$, yielding a highly
sparse linear equation
\begin{equation}
\left(\mtbfL+\alpha\mtbfI\right)\mtbfh=\alpha\mtbfh^{\left(0\right)},\label{eq:highlySparseLinearEquation}
\end{equation}
 which could be efficiently solved~\cite{ALevin_2008_PAMI_closedFormd}.
In order to simplify the incorporation of the learnt DgMap into the
$\ell_{p}\!\left(0\!<\! p\!<\!1\right)$ norm, the data-guided values
are resized into the range of $\left(0,1\right)$ as 
\[
h_{n}\leftarrow\frac{h_{n}-\min\left(\mtbfh\right)}{\max\left(\mtbfh\right)-\min\left(\mtbfh\right)+\beta},\quad n=1,2,\cdots,N,
\]
 where $\beta\!=\!10^{-8}$ is a small value used to prevent $\left\{ h_{n}\right\} _{n=1}^{N}$
being equal to $1$, ensuring the numerical stability for the sparse
constraint in the next section.

To study the influence of the local window size, we conduct an experiment
as illustrated in Fig.~\ref{fig:fineTunedDgMap}, where Fig.~\ref{fig:fineTunedDgMap-1}
shows the reference \emph{abundance} map, Fig.~\ref{fig:fineTunedDgMap-2}
shows the initial DgMap, Fig.~\ref{fig:fineTunedDgMap-3} shows the
refined DgMap with $3\times3$ local window, followed by the fine
tuned result with $7\times7$ local window in Fig.~\ref{fig:fineTunedDgMap-4}.
As Fig.~\ref{fig:fineTunedDgMap} shows, the $3\times3$ local window
is sufficient to get suitable result at low computational costs. Therefore,
the $3\times3$ local window size is chosen in this work.

\section{\label{sec:Data-guided-Sparse-NMF}Data-guided Sparse NMF (DgS-NMF)}

\subsection{Data-guided  Regularization and DgS-NMF Model }

Based on the linear combination model in~\eqref{eq:linearMixtureModel},
a hyperspectral image $\mtbfY\!\triangleq\!\left[\mtbfy_{1},\mtbfy_{2},\cdots,\mtbfy_{N}\right]\in\mtbbR_{+}^{L\times N}$,
with $L$ channels and $N$ pixels, could be approximated by two factor
matrices:
\begin{figure}[tb]
\centering{}\includegraphics[width=0.82\columnwidth]{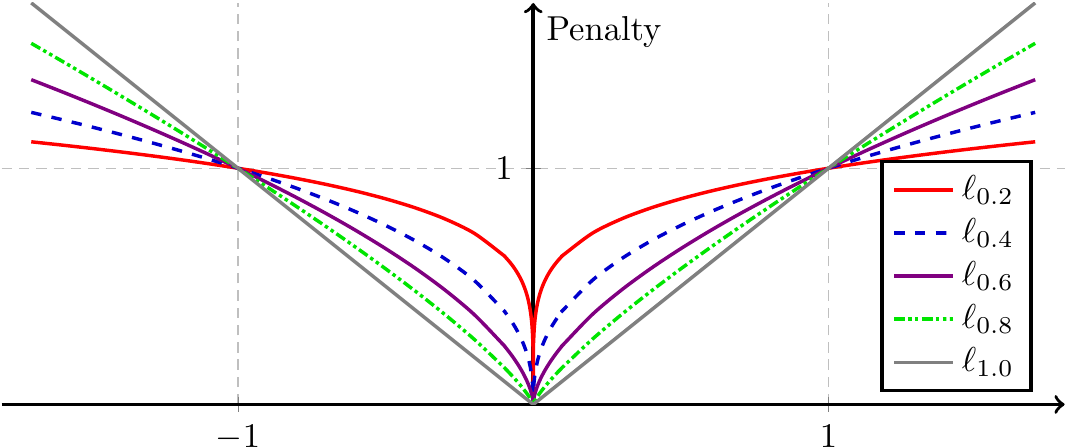}\caption{The shape of the $\ell_{p}$-norm with different $p\in(0,1]$, indicating
that a smaller $p$ tends to find a sparser solution~\cite{Fan_2001_JASA_LaNorm}.
\label{fig:sparseCurves_2_Lomega}}
\end{figure}
 
\begin{equation}
\mtbfY=\mtbfM\mtbfA+\mtbfE,\label{eq:linearMixtureModel_matrix}
\end{equation}
where $\mtbfM\!\triangleq\!\left[\mtbfm_{1},\cdots,\mtbfm_{K}\right]\!\in\!\mtbbR_{+}^{L\times K}$
is the \emph{endmember} matrix including $K$ spectral vectors, $K\!\ll\!\min\!\left\{ L,N\right\} $;
$\mtbfA\!\triangleq\!\left[\mtbfa_{1},\cdots,\mtbfa_{N}\right]\!\in\!\mtbbR_{+}^{K\times N}$
is the corresponding\emph{ abundance} matrix, whose $n^{\mtth}$ column
vector $\mtbfa_{n}$ contains all the $K$ \emph{abundances} at pixel
$\mtbfy_{n}$; $\mtbfE$ is a residual term. Specifically,~\eqref{eq:linearMixtureModel_matrix}
could be naturally translated into the Nonnegative Matrix Factorization~\cite{Lee_99_Nature_NMF}
(NMF) problem by strictly constraining the nonnegative property of
both factors, i.e. $\mtbfM\!\geq\!\mtbfzero,\mtbfA\!\geq\!\mtbfzero$,
which agrees with the nonnegative requirement on both\emph{ endmembers}
and \emph{abundances}. Such case suggests that NMF is physically suitable
for the HU task.

Suppose we are given the fine tuned DgMap $\mtbfh\!\in\!\mtbbR_{+}^{N}$.
Different from the traditional sparse regularization~\cite{Hoyer_02_NNSP_NMF_l1,Qian_11_TGRS_NMF+l1/2}
that constrains all factors $\left\{ \mtbfa_{n}\right\} _{n=1}^{N}$
at the same sparse level as 
\begin{equation}
\mtcalJ\left(\mtbfA\right)=\sum_{n=1}^{N}\left\Vert \mtbfa_{n}\right\Vert _{1},\quad\mtcalJ\left(\mtbfA\right)=\sum_{n=1}^{N}\left\Vert \mtbfa_{n}\right\Vert _{1/2}^{1/2},\label{eq:traditionalSprsConstaint}
\end{equation}
where $\left\Vert \mtbfa\right\Vert _{p}^{p}=\sum_{k}\left|a_{k}\right|^{p}\left(\forall0<p<1\right)$,
this paper proposes a novel data-guided constraint as 
\begin{align}
\mtcalJ\left(\mtbfA\right) & =\sum_{n=1}^{N}\left\Vert \mtbfa_{n}\right\Vert _{1-h_{n}}^{1-h_{n}}=\sum_{n=1}^{N}\left(\sum_{k=1}^{K}\left|A_{kn}\right|^{1-H_{kn}}\right),\label{eq:ourSparseConstaint}
\end{align}
where $H_{kn}$ is the $\left(k,n\right)$-th element in the matrix
$\mtbfH\!=\!\mtbfone_{K}\mtbfh^{T}$. The elements in the same column
of $\mtbfH$ are identical to each other, i.e. $H_{1,n}\!=\! H_{2,n}\cdots=\! H_{K,n}\!=\! h_{n},\forall n\!\in\!\left\{ 1,\cdots,N\right\} $. 

In this way, all the \emph{abundance} factors $\left\{ \mtbfa_{n}\right\} _{n=1}^{N}$
are constrained in the $\ell_{p}\left(0\!<\! p\!<\!1\right)$-norm.
For each factor, the level (strength) of sparse constraint is closely
related to the choice of $p$---a smaller $p$ corresponds to a sparser
constraint~\cite{Fan_2001_JASA_LaNorm} (cf. Fig.~\ref{fig:sparseCurves_2_Lomega}).
This amounts to the dependence on the DgMap value $h_{n}$, as shown
in~\eqref{eq:ourSparseConstaint}. So, for instance, a pixel for
which $h_{n}=0.2$ will be constrained by a weak sparsity regularization
in the $\ell_{0.8}$-norm, whereas one for which $h_{n}=0.8$ will
be constrained by the $\ell_{0.2}$ regularization and so will enjoy
a heavy sparsity constraint (cf. Fig.~\ref{fig:sparseCurves_2_Lomega}).
Additionally, the DgMap values in the transition areas are generally
small (cf. Fig.~\ref{fig:fineTunedDgMap-3}). As a result, they will
be constrained at relatively low levels of sparsity constraints, conforming
to their mixed properties.

Compared with the traditional regularization~\eqref{eq:traditionalSprsConstaint},
the advantages of our constraint~\eqref{eq:ourSparseConstaint} lie
in three aspects: 1) as the fine tuned DgMap describes the mixed level
over the entire image, our constraint is more agreeable with the practical
mixed property of each pixel; 2) with the careful constraint in~\eqref{eq:ourSparseConstaint},
the non-convex objective function~\eqref{eq:objectiveFunction_DgS-nmf}
is more likely to converge to some suitable local minima; 3) although
the adaptive sparsity regularization is constrained on the \emph{abundance}
factors, it would explicitly influence the estimation of \emph{endmembers},
guiding the \emph{endmembers} toward the pixels with highly sparse
constraint. This doesn't mean that the pixels with highly sparse constraints
are \emph{endmembers}. Many pixels with highly sparse constraints
compete for the \emph{endmember, }and some\emph{ }trade-off spectra
could also be \emph{endmembers.}

Apart from the advantages above, it is easy to find that the traditional
sparse constraints~\eqref{eq:traditionalSprsConstaint} are special
cases of our adaptive sparse constraint~\eqref{eq:ourSparseConstaint}.
Given a constant DgMap with each pixel $\left\{ h_{n}\right\} _{n=1}^{N}$
equal to zero, the adaptive sparse constraint degrades into the $\ell_{1}$
regularization, i.e. $\mtcalJ\left(\mtbfA\right)=\sum_{n}\left\Vert \mtbfa_{n}\right\Vert _{1-h_{n}}^{1-h_{n}}=\sum_{n}\left\Vert \mtbfa_{n}\right\Vert _{1}$;
whereas if each element in the DgMap is equal to $1/2$, the adaptive
sparse constraint turns into the $\ell_{1/2}$ regularization. Moreover,
for the HU task, all the elements in $\mtbfA$ are within the range
of $\left(0,1\right)$~\cite{LiuXueSong_2011_TGRS_ConstrainedNMF,Qian_11_TGRS_NMF+l1/2}.
Thus, once allowed the limit $h_{n}\rightarrow-\infty$, the adaptive
sparse constraint would degrade into the non-regularization case,
i.e. $\mtcalJ\left(\mtbfA\right)\rightarrow0$, since for any $a\in\left(0,1\right)$,
we have $a^{1+\infty}\rightarrow0$.

To obtain the optimal factor matrices, we model the matrix representation
problem~\eqref{eq:linearMixtureModel_matrix} as the Data-guided
Sparsity regularized Nonnegative Matrix\textbf{ }Factorization (DgS-NMF)
objective
\begin{align}
\mtcalO\left(\mtbfM,\mtbfA\right) & =\frac{1}{2}\left\Vert \mtbfY-\mtbfM\mtbfA\right\Vert _{F}^{2}+\lambda\sum_{n=1}^{N}\sum_{k=1}^{K}\left|A_{kn}\right|^{1-H_{kn}}\label{eq:objectiveFunction_DgS-nmf}\\
 & \mtsubTo\ \mtbfM\geq\mtbfzero,\mtbfA\geq\mtbfzero,\nonumber 
\end{align}
where $\lambda\geq0$ is a balancing parameter that controls the average
sparsity of the factor representation. In the next subsection, the
optimization for the DgS-NMF problem will be analyzed.

\subsection{Updating Rules  for DgS-NMF}

Akin to NMF~\cite{Lee_00_NIPS_NMF} and EM\ \cite{Dempster_77_JRSS_EM},
the objective function in~\eqref{eq:objectiveFunction_DgS-nmf} is
non-convex for $\mtbfM$ and $\mtbfA$ together. No global minima
could be reached. Alternatively, we propose an iterative algorithm
that alternately updates $\mtbfM$ and $\mtbfA$ at each iteration.
It has the ability to arrive at some local minima after finite iterations,
which will be proved in Section\ \ref{sub:ConvergenceProof_DgS-nmf}. 

Specifically, the Lipschitz constant\ \cite{JChYe_2012_JNM_Lipschitz}
of the data-guided constraint\ \eqref{eq:ourSparseConstaint} will
be infinity for $A_{kn}=0,\forall k,n$. To ensure the Lipschitz condition,
we reformulate our model\ \eqref{eq:objectiveFunction_DgS-nmf} as
\begin{align}
\mtcalO\left(\mtbfM,\mtbfA\right) & =\frac{1}{2}\left\Vert \mtbfY-\mtbfM\mtbfA\right\Vert _{F}^{2}+\lambda\sum_{n=1}^{N}\sum_{k=1}^{K}\left(A_{kn}+\xi\right)^{1-H_{kn}}\nonumber \\
 & \mtsubTo\ \mtbfM\geq\mtbfzero,\mtbfA\geq\mtbfzero,\label{eq:objectiveFunction_DgS_nmf_Lipschitz}
\end{align}
where $\xi$ is a small positive value to ensure the numerical condition.
It is obvious that the objective\ \eqref{eq:objectiveFunction_DgS_nmf_Lipschitz}
is reduced to\ \eqref{eq:objectiveFunction_DgS-nmf} when $\xi\rightarrow0$.
For simplicity, we use $\mtbfA+\xi\!=\!\left[A_{kn}+\xi\right]$ to
expresses the idea of adding $\xi$ to every entry $A_{kn},\forall k,n$.

Considering the constraints of $\mtbfM\!\geq\!\mtbfzero,\mtbfA\!\geq\!\mtbfzero$,
the objective function~\eqref{eq:objectiveFunction_DgS_nmf_Lipschitz}
could be rewritten as the Lagrange Multiplier: 
\begin{align}
\mtcalL= & \frac{1}{2}\left\Vert \mtbfY-\mtbfM\mtbfA\right\Vert _{F}^{2}+\lambda\sum_{n=1}^{N}\sum_{k=1}^{K}\left(A_{kn}+\xi\right)^{1-H_{kn}}\nonumber \\
 & +\mttrace\left(\Psi\mtbfM^{T}\right)+\mttrace\left(\Gamma\mtbfA^{T}\right),\label{eq:LaplacianMultiplier_DgS-nmf}
\end{align}
where $\psi_{lk},\gamma_{kn}$ are the lagrange multipliers for the
inequality constraints $M_{lk}\!\geq\!0$ and $A_{kn}\!\geq\!0$ respectively,
and $\Psi\!=\!\left[\psi_{lk}\right]\!\in\!\mtbbR_{+}^{L\times K}$,
$\Gamma\!=\!\left[\gamma_{kn}\right]\!\in\!\mtbbR_{+}^{K\times N}$
are the lagrange multipliers in matrix format. To find the local minima,
one intuitive approach is to differentiate~\eqref{eq:LaplacianMultiplier_DgS-nmf}
and set the partial derivatives to zero. This amounts to solving the
following linear equations
\begin{equation}
\nabla_{\mtbfM}\mtcalL=\mtbfM\mtbfA\mtbfA^{T}-\mtbfY\mtbfA^{T}+\Psi=\mtbfzero\label{eq:partialDerive_M}
\end{equation}
\begin{align}
\nabla_{\mtbfA}\mtcalL= & \mtbfM^{T}\mtbfM\mtbfA-\mtbfM^{T}\mtbfY+\Gamma+\nonumber \\
 & \lambda\left(\mtbfone-\mtbfH\right)\circ\left(\mtbfA+\xi\right)^{-\mtbfH}=\mtbfzero,\label{eq:partialDerive_A}
\end{align}
where $\circ$ is the Hadamard product between matrices; $\mtbfA^{\mtbfH}\!=\!\left[\left(A_{kn}\right)^{H_{kn}}\right]\!\in\!\mtbbR_{+}^{K\times N}$
is an elementwise exponential operation. Based on the Karush-Kuhn-Tucker
conditions $\psi_{lk}M_{lk}\!=\!0$ and $\gamma_{kn}A_{kn}\!=\!0$,
we could simplify~\eqref{eq:partialDerive_M} and~\eqref{eq:partialDerive_A}
by multiplying both sides with $M_{lk}$ and $A_{kn}$ respectively,
yielding 
\begin{align}
\left(\mtbfM\mtbfA\mtbfA^{T}\right)_{lk}M_{lk}-\left(\mtbfY\mtbfA^{T}\right)_{lk}M_{lk} & =0\label{eq:partDeriveEquaM}\\
\left(\mtbfM^{T}\mtbfM\mtbfA\right)_{kn}A_{kn}-\left(\mtbfM^{T}\mtbfY\right)_{kn}A_{kn}+\nonumber \\
\lambda\left(\left(\mtbfone-\mtbfH\right)\circ\left(\mtbfA+\xi\right)^{-\mtbfH}\right)_{kn}A_{kn} & =0.\label{eq:partDeriveEquaA}
\end{align}
Solving Eqs.~\eqref{eq:partDeriveEquaM} and~\eqref{eq:partDeriveEquaA},
we get the updating rules as
\begin{align}
M_{lk} & \leftarrow M_{lk}\frac{\left(\mtbfY\mtbfA^{T}\right)_{lk}}{\left(\mtbfM\mtbfA\mtbfA^{T}\right)_{lk}}\label{eq:updataDgS_M}\\
A_{kn} & \leftarrow A_{kn}\frac{\left(\mtbfM^{T}\mtbfY\right)_{kn}}{\left(\mtbfM^{T}\mtbfM\mtbfA+\lambda\left(\mtbfone-\mtbfH\right)\circ\left(\mtbfA+\xi\right)^{-\mtbfH}\right)_{kn}}.\label{eq:updateDgS_A}
\end{align}

\begin{algorithm}[t]
\caption{\textbf{for DgS-NMF \label{alg:DgS_nmf}}}
 \textbf{Input:} the hyperspectral image $\mtbfY\!\in\!\mtbbR_{+}^{L\times N}$,
the number of \emph{endmembers} (i.e. $K$) and the penalty parameters
$\lambda$. \\
\textbf{Output:} two factor matrices $\mtbfM\!\in\!\mtbbR_{+}^{L\times K}$
and $\mtbfA\!\in\!\mtbbR_{+}^{K\times N}$.

\begin{algorithmic}[1] 

\STATE Calculate initial DgMap $\mtbfh^{\left(0\right)}\!\in\!\mtbbR_{+}^{N}$
according to  Eq.\ \eqref{eq:initialDgMap}. 

\STATE Get the fine tuned DgMap $\mtbfh$ by solving the highly sparse
linear equation\ \eqref{eq:highlySparseLinearEquation}. Calculate
$\mtbfH=\mtbfone_{K}\mtbfh^{T}\in\mtbbR^{K\times N}.$ 

\STATE Initialize\emph{ }the factor matrices $\mtbfM$ and $\mtbfA$.

\REPEAT

\STATE update $\mtbfA$ by the updating rule~\eqref{eq:updateDgS_A}.

\STATE update $\mtbfM$ by the updating rule~\eqref{eq:updataDgS_M}.

\STATE scale $\mtbfM$ and $\mtbfA$ by Eq.~\eqref{eq:eleminatingUncertain}
after each iteration.

\UNTIL{convergence} \\
\STATE Output $\mtbfM$ and $\mtbfA$ as the final unmixing result. 

\end{algorithmic} 
\end{algorithm}
However, if $\mtbfM$ and $\mtbfA$ form the solution of NMF, $\mtbfD\mtbfU$
and $\mtbfU^{-1}\mtbfA$ are the solution for any positive diagonal
matrix $\mtbfU$~\cite{Cai_11_PAMI_GNMF,fyzhu_2014_IJPRS_SS_NMF}.
To get rid of this kind of uncertainty, one intuitive method is to
scale each row of $\mtbfA$ or each column of $\mtbfM$ to be unit
$\ell_{1}$-norm or $\ell_{2}$-norm~\cite{WeiXu_03_SIGIR_DocClusterNMF}
as follows 
\begin{equation}
M_{lk}\leftarrow M_{lk}\left(\sum_{n=1}^{N}\left|A_{kn}\right|\right),\quad A_{kn}\leftarrow\frac{A_{kn}}{\sum_{n=1}^{N}\left|A_{kn}\right|}.\label{eq:eleminatingUncertain}
\end{equation}
Similarly, we scale $\mtbfM$ and $\mtbfA$ by\ \eqref{eq:eleminatingUncertain}
after each iteration.

The algorithm for DgS-NMF is summarized in Algorithm~\ref{alg:DgS_nmf}.
For the updating rules in~\eqref{eq:updataDgS_M} and~\eqref{eq:updateDgS_A},
we have the following theorem, which will be proven in the next section,
as 
\begin{thm}
\label{thm:nonincreasing_theorem}The objective function~\eqref{eq:objectiveFunction_DgS-nmf}
is non-increasing under the updating rules~\eqref{eq:updataDgS_M}
and~\eqref{eq:updateDgS_A}. 
\end{thm}

\subsection{\label{sub:ConvergenceProof_DgS-nmf}Convergence Proof for DgS-NMF}

To ensure the reliability of~\eqref{eq:updataDgS_M} and~\eqref{eq:updateDgS_A},
the convergence proofs of both updating rules are discussed. Fortunately,
the convergence proof of~\eqref{eq:updataDgS_M} could be eliminated
since it has been analyzed in~\cite{Lee_00_NIPS_NMF}. A common skill
used in EM~\cite{Dempster_77_JRSS_EM,Saul_97_EMNLP_mixedOrderMarkov}
and NMF\ \cite{Lee_00_NIPS_NMF} is employed by introducing an auxiliary
function:
\begin{defn}
$G\left(\mtbfA,\mtbfA'\right)$ is an auxiliary function of $\mtcalO\left(\mtbfA\right)$
if the following properties are satisfied, 
\begin{equation}
G\left(\mtbfA,\mtbfA'\right)\geq\mtcalO\left(\mtbfA\right),\quad G\left(\mtbfA,\mtbfA\right)=\mtcalO\left(\mtbfA\right).
\end{equation}
\end{defn}
\begin{lem}
\label{thm:convergenceWithAuxiliaryFunction} By minimizing the energy
of $G\left(\mtbfA,\mtbfA'\right)$ given by 
\[
\mtbfA^{\left(t+1\right)}=\arg\min_{\mtbfA}G\left(\mtbfA,\mtbfA^{\left(t\right)}\right),
\]
we can obtain a solution $\mtbfA^{\left(t+1\right)}$ that makes $\mtcalO\left(\mtbfA\right)$
non-increasing at each iteration, i.e. $\mtcalO\left(\mtbfA^{\left(t+1\right)}\right)\leq\mtcalO\left(\mtbfA^{\left(t\right)}\right).$
Finally, $\mtcalO\left(\mtbfA\right)$ will converge after finite
iterations. \end{lem}
\begin{IEEEproof}
This is because of the following inequalities:
\begin{align*}
\mtcalO\left(\mtbfA^{\left(\text{min}\right)}\right) & \leq\cdots\leq\mtcalO\left(\mtbfA^{\left(t+1\right)}\right)\leq G\left(\mtbfA^{\left(t+1\right)},\mtbfA^{\left(t\right)}\right)\\
 & \leq\mtcalO\left(\mtbfA^{\left(t\right)}\right)\leq\cdots\leq\mtcalO\left(\mtbfA^{\left(\text{0}\right)}\right)
\end{align*}

\end{IEEEproof}
Now we consider the objective function~\eqref{eq:objectiveFunction_DgS_nmf_Lipschitz}
with $\mtbfA$ as the only variable:
\begin{equation}
\mtcalO\left(\mtbfA\right)=\frac{1}{2}\|\mtbfY-\mtbfM\mtbfA\|_{F}^{2}+\lambda\sum_{n=1}^{N}\sum_{k=1}^{K}\left(A_{kn}^{\left(t\right)}+\xi\right)^{1-H_{kn}}.\label{eq:O_func_A}
\end{equation}
Specifically, it is approximately a quadratic function as follows
\begin{align}
\mtcalO\left(\mtbfA\right)\approx & \mtcalO\left(\mtbfA^{\left(t\right)}\right)+\mttrace\left(\mtbfC^{T}\nabla\mtcalO\left(\mtbfA^{\left(t\right)}\right)\right)\nonumber \\
 & +\frac{1}{2}\left[\mttrace\left(\mtbfC^{T}\left(\mtbfM^{T}\mtbfM\right)\mtbfC\right)-\lambda F\left(\mtbfA\right)\right],\label{eq:objTaylorExpan}
\end{align}
where $\mtbfC=\left(\mtbfA-\mtbfA^{\left(t\right)}\right)$ and 
\[
F\left(\mtbfA\right)=\sum_{n,k}H_{kn}\left(1-H_{kn}\right)\left(A_{kn}^{\left(t\right)}+\xi\right)^{-\left(H_{kn}+1\right)}C_{kn}^{2}.
\]
To prove the convergence property of~\eqref{eq:updateDgS_A}, we
have to find an auxiliary function of~\eqref{eq:objTaylorExpan},
by which the updating rule~\eqref{eq:updateDgS_A} could be obtained
by differentiating this auxiliary function and setting the derivatives
to zero. Conversely, a function constituted based on the updating
rule~\eqref{eq:updateDgS_A} is given by
\begin{align}
G\left(\mtbfA,\mtbfA^{\left(t\right)}\right)= & \mtcalO\left(\mtbfA^{\left(t\right)}\right)+\mttrace\left(\mtbfC^{T}\nabla\mtcalO\left(\mtbfA^{\left(t\right)}\right)\right)\nonumber \\
 & +\frac{1}{2}\sum_{k=1}^{K}\sum_{n=1}^{N}Q_{kn}C_{kn}^{2},\label{eq:auxiliaryFunc}
\end{align}
where
\[
Q_{kn}=\frac{\left(\mtbfM^{T}\mtbfM\mtbfA^{\left(t\right)}+\lambda\left(\mtbfone-\mtbfH\right)\circ\left(\mtbfA^{\left(t\right)}+\xi\right)^{-\mtbfH}\right)_{kn}}{A_{kn}^{\left(t\right)}}.
\]
 It could be separated into two parts $Q_{kn}=Q_{kn}^{\left(1\right)}+\lambda Q_{kn}^{\left(2\right)}$
as
\[
Q_{kn}^{\left(1\right)}=\left(\sum_{l=1}^{K}\frac{\left(\mtbfM^{T}\mtbfM\right)_{kl}A_{ln}^{\left(t\right)}}{A_{kn}^{\left(t\right)}}\right)
\]
\[
Q_{kn}^{\left(2\right)}=\frac{\left(1-H_{kn}\right)\left(A_{kn}^{\left(t\right)}+\xi\right)^{-H_{kn}}}{A_{kn}^{\left(t\right)}}.
\]
 Since $\frac{\left(A_{kn}^{\left(t\right)}+\xi\right)^{-H_{kn}}}{A_{kn}^{\left(t\right)}}>\left(A_{kn}^{\left(t\right)}+\xi\right)^{-\left(H_{kn}+1\right)}$,
we have 
\begin{align}
Q_{kn}^{\left(2\right)} & >\left(1-H_{kn}\right)\left(A_{kn}^{\left(t\right)}+\xi\right)^{-\left(H_{kn}+1\right)}.\label{eq:new_inequalityProof}
\end{align}
 Specifically, we have to prove the following lemma:
\begin{lem}
\label{lem:proofUpperFunct}The function $G\left(\mtbfA,\mtbfA^{\left(t\right)}\right)$
defined in~\eqref{eq:auxiliaryFunc} is an auxiliary function for
$\mtcalO\left(\mtbfA\right)$ defined in~\eqref{eq:objTaylorExpan}. \end{lem}
\begin{IEEEproof}
On the one hand, the equation $\mtcalO\left(\mtbfA\right)=\mtcalO\left(\mtbfA^{\left(t\right)}\right)=G\left(\mtbfA,\mtbfA^{\left(t\right)}\right)$
holds for any $\mtbfA=\mtbfA^{\left(t\right)}$, i.e. $\mtbfC=\mtbfzero$.
On the other hand, when $\mtbfA\neq\mtbfA^{\left(t\right)}$, i.e.
$\mtbfC\neq\mtbfzero$ we have to prove $\mtcalO\left(\mtbfA^{\left(t\right)}\right)\leq G\left(\mtbfA,\mtbfA^{\left(t\right)}\right)$.

Since the constant term and linear term in~\eqref{eq:objTaylorExpan}
are identical to their counterparts in~\eqref{eq:auxiliaryFunc},
Lemma~\ref{lem:proofUpperFunct} could be proven by only comparing
the quadratic terms as
\begin{equation}
\sum_{k=1}^{K}\sum_{n=1}^{N}Q_{kn}C_{kn}^{2}\geq\mttrace\left(\mtbfC^{T}\left(\mtbfM^{T}\mtbfM\right)\mtbfC\right)-F\left(\mtbfA\right).\label{eq:auxQuadraticTermNew}
\end{equation}
The inequality above could be expressed as two terms 
\begin{equation}
\underset{\text{first term}}{\underbrace{\sum_{k,n}Q_{kn}^{\left(1\right)}C_{kn}^{2}-\text{Tr}\left(\mtbfC^{T}\mtbfM^{T}\mtbfM\mtbfC\right)}}+\lambda\underset{\text{second term}}{\underbrace{\left(\sum_{k,n}Q_{kn}^{\left(2\right)}C_{kn}^{2}+F\left(\mtbfA\right)\right)}\geq0.}\label{eq:compareAuxAndTaylor_v2}
\end{equation}
We could prove the inequality\ \eqref{eq:compareAuxAndTaylor_v2}
by verifying that both terms are greater than or equal to zero. Therefore,
the inequality~\eqref{eq:compareAuxAndTaylor_v2} could be proven
by comparing the first term\ \cite{fyzhu_2014_IJPRS_SS_NMF} 
\begin{align}
f_{1}= & \sum_{k,n,l}\left(\frac{\left(\mtbfM^{T}\mtbfM\right)_{kl}A_{ln}^{\left(t\right)}}{A_{kn}^{\left(t\right)}}C_{kn}^{2}-C_{kn}C_{ln}\left(\mtbfM^{T}\mtbfM\right)_{lk}\right)\nonumber \\
= & \sum_{k,n,l}\frac{\left(\mtbfM^{T}\mtbfM\right)_{kl}}{2A_{kn}^{\left(t\right)}A_{ln}^{\left(t\right)}}\left(A_{ln}^{\left(t\right)}C_{kn}-A_{kn}^{\left(t\right)}C_{ln}\right)^{2}\geq0.\label{eq:auxProofCha_1}
\end{align}
Then considering the inequality\ \eqref{eq:new_inequalityProof},
the second term becomes 
\begin{align}
f_{2}> & \sum_{k,n}\left(1-H_{kn}\right)\left(A_{kn}^{\left(t\right)}+\xi\right)^{-\left(H_{kn}+1\right)}C_{kn}^{2}+\nonumber \\
 & \sum_{k,n}H_{kn}\left(1-H_{kn}\right)\left(A_{kn}^{\left(t\right)}+\xi\right)^{-\left(H_{kn}+1\right)}C_{kn}^{2}\nonumber \\
= & \sum_{k,n}\left(1-H_{kn}^{2}\right)\left(A_{kn}^{\left(t\right)}+\xi\right)^{-\left(H_{kn}+1\right)}C_{kn}^{2},\label{eq:auxProofCha_2}
\end{align}
 where $C_{kn}^{2}$ is undoubtedly nonnegative. Since any element
$H_{kn}$ lies in the range of $\left(0,1\right)$, this ensures the
nonnegative property of $(1-H_{kn}^{2})$. The expression $(A_{kn}^{\left(t\right)}+\xi)^{-\left(H_{kn}+1\right)}$
is positive as $A_{kn}^{\left(t\right)}+\xi$ is always positive.
Therefore, $f_{2}\geq0$ holds for any condition. We have proven the
inequality~\eqref{eq:auxQuadraticTermNew} or\ \eqref{eq:compareAuxAndTaylor_v2}
by proving $f_{1}\geq0$ and $f_{2}\geq0$. In consequence, $G\left(\mtbfA,\mtbfA^{\left(t\right)}\right)$
is an auxiliary function of $\mtcalO\left(\mtbfA\right)$. 
\end{IEEEproof}
Through the theoretical analyses above, we have proven Theorem~\ref{thm:nonincreasing_theorem}.
In addition, the empirical convergence property of DgS-NMF will be
analyzed in Section~\ref{sub:Convergence-Study}. 
\begin{figure*}[tb]
\begin{centering}
\scalebox{0.95}{\subfloat[{\small{Samson }}\label{fig:realHyperImages_samson}]{\begin{centering}
\includegraphics[width=0.32\columnwidth]{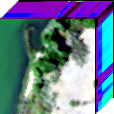}
\par\end{centering}

\centering{}}\subfloat[{\small{Jasper Ridge }}\label{fig:realHyperImages_japser}]{\begin{centering}
\includegraphics[width=0.35\columnwidth]{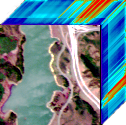}
\par\end{centering}

\centering{}}\subfloat[{\small{Urban }}\label{fig:realHyperImages_urban}]{\begin{centering}
\includegraphics[width=0.48\columnwidth]{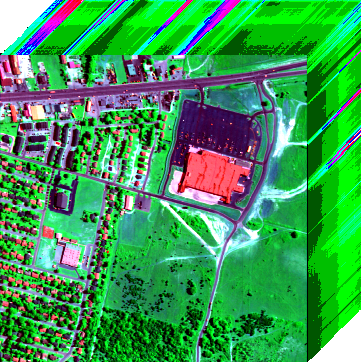}
\par\end{centering}

\centering{}}\subfloat[{\small{Cuprite }}\label{fig:realHyperImages_cuprite}{\small{ }}]{\begin{centering}
\includegraphics[width=0.35\columnwidth]{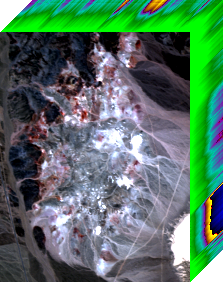}
\par\end{centering}

\centering{}}}
\par\end{centering}

\centering{}\caption{The four real hyperspectral images, i.e. Samson, Jasper Ridge, Urban
and Cuprite respectively, used in the experiments.  \label{fig:realHyperImages_DgMaps}}
\end{figure*}
\begin{table*}[tb]
\centering{}\caption{Computational operation counts for NMF and {\small{DgS-NMF}} at each
iteration.\label{tab:ComplexSummarize}}
\begin{tabular}{|c||c|c|c|c||c|}
\hline 
\multirow{2}{*}{Methods} &
\multicolumn{4}{c||}{Arithmetic Operations in float-point format} &
\multirow{2}{*}{Overall}\tabularnewline
\cline{2-5} 
 & Addition &
Multiplication &
Division &
Exponent & \tabularnewline
\hline 
\multirow{2}{*}{NMF} &
$2LNK+2K^{2}\left(L+N\right)$ &
$2LNK+2K^{2}\left(L+N\right)$ &
\multirow{2}{*}{$K\left(L+N\right)$} &
\multirow{2}{*}{--} &
\multirow{2}{*}{$O\left(KLN\right)$}\tabularnewline
 & $-2K\left(N+L\right)-2K^{2}$ &
$+K\left(L+N\right)$ &  &  & \tabularnewline
\hline 
\multirow{2}{*}{DgS-NMF} &
$2LNK+2K^{2}\left(L+N\right)$ &
$2LNK+2K^{2}\left(L+N\right)$ &
\multirow{2}{*}{$K\left(L+N\right)$} &
\multirow{2}{*}{$KN$} &
\multirow{2}{*}{$O\left(KLN\right)$}\tabularnewline
 & $-K\left(N+2L\right)-2K^{2}$ &
$+K\left(L+3N\right)$ &  &  & \tabularnewline
\hline 
\end{tabular}
\end{table*}
\begin{table}[t]
\caption{Parameters used in Computational Complexity Analysis. \label{tab:complexParameters}}

\centering{}%
\begin{tabular}{|c|l|}
\hline 
Parameters &
Description\tabularnewline
\hline 
$K$ &
number of \emph{endmembers}\tabularnewline
\hline 
 $L$ &
number of channels\tabularnewline
\hline 
$N$ &
number of pixels in hyperspectral image\tabularnewline
\hline 
$t$ &
number of iterations\tabularnewline
\hline 
$q\left(=9\right)$ &
number of pixels in the local window\tabularnewline
\hline 
\end{tabular}\vspace{-0.15cm}
\end{table}

\subsection{\label{sub:Computational-Complexity-Analysi}Computational Complexity
Analysis for DgS-NMF }

Speed is important for algorithms\ \cite{Nocedal_2006_book_numericalOptimization,fyzhu_2014_AAAI_ARSS}.
For this reason, the computational complexity of DgS-NMF is thoroughly
analyzed by comparing with that of NMF. Since both algorithms are
iteratively updated, the complexity is analyzed by summarizing the
arithmetic operations at each iteration, then considering the iteration
steps. For convenience, the parameters used here are listed in Table~\ref{tab:complexParameters}.

In the updating rules~\eqref{eq:updataDgS_M} and~\eqref{eq:updateDgS_A},
there are four kinds of arithmetic operations, i.e. addition, multiplication,
division and exponent respectively. Table~\ref{tab:ComplexSummarize}
summaries the counts of each arithmetic operation as well as the overall
cost. In terms of the four operations, the differences between DgS-NMF
and NMF are limited: DgS-NMF requires $KN$ more additions, $2KN$
more multiplications and $KN$ more exponents. Nevertheless, both
methods have a $O\left(KLN\right)$ overall cost at one iteration
step, as shown in the last column of Table~\ref{tab:ComplexSummarize}.

Apart from the updating costs, the DgS-NMF method requires $O\left(qLN\right)$
to obtain the initial DgMap and $O\left(q^{2}LN+\left(2\sqrt{q}-1\right)^{2}N\right)$~\cite{Davis_2006_SIAM_SparseLinearSystems,Bishop_06_Springer_PRML,matlabSprsLinearSystemComplexity}
to get the fine tuned one. Thus, if both methods needs $t$ iterations,
the total computational complexities are $O\left(tKLN\right)$ for
NMF and
\[
O\left(tKLN+qLN+q^{2}LN+\left(2\sqrt{q}-1\right)^{2}N\right)
\]
 for DgS-NMF. For the HU task, we have $N\gg\max\left(K,L,t,q\right)$,
thus, indicating that the computational complexity of DgS-NMF is a
only bit more than that of NMF, but still in the same order of magnitude.

\section{\label{sec:Evaluation}Evaluation}

In this section, we evaluate the performance of the proposed method
for the HU task. Several experiments are carried out to show that
DgS-NMF is successfully adapted to the HU task. 
\begin{figure*}[t]
\begin{centering}
\includegraphics[width=2.05\columnwidth]{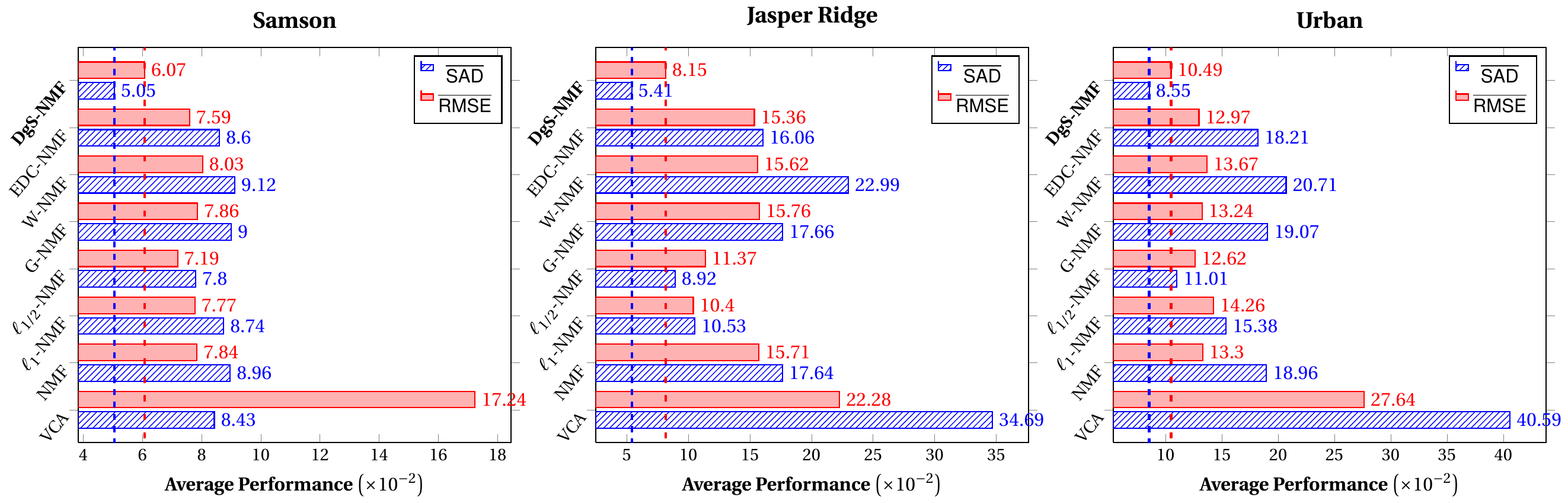}
\par\end{centering}

\centering{}\caption{The average performances (i.e. $\overline{\text{SAD}}$ and $\overline{\text{RMSE}}$)
of eight methods on the three datasets: Samson, Jasper Ridge and Urban,
respectively. \label{fig:AveragedPerformanceOn3DataSets}}
\end{figure*}

\subsection{Real Hyperspectral Images}

This section introduces the information of four hyperspectral data
used in the experiment. Specifically, the ground truth is achieved
via the method introduced in\ \cite{fyzhu_2014_GroundTruth4HU,fyzhu_2014_JSTSP_RRLbS,Jia_09_TGRS_ConstainedNMF}.

\textbf{Samson}, as shown in Fig.~\ref{fig:realHyperImages_samson},
is an simple data available on\ \url{http://opticks.org/confluence/display/opticks/Sample+Data}.
There are $952\!\times\!952$ pixels in it. Each pixel is observed
at $156$ channels covering the wavelength from $0.401$ to $0.889\mu m$.
As a result, the spectral resolution is highly up to $3.13nm$. The
original image is very large, which could be computationally expensive
for the HU study. A region of $95\!\times\!95$ pixels is considered,
whose first pixel corresponds to the $\left(252,332\right)$-th pixel
in the original image. There are three \emph{endmembers} in this image,
i.e. `\#1 Soil', `\#2 Tree' and `\#3 Water'.

\textbf{Jasper Ridge},\textbf{ }as shown in Fig.~\ref{fig:realHyperImages_japser},\textbf{
}is a popular hyperspectral data used in~\cite{enviTutorials,fyzhu_2014_IJPRS_SS_NMF}.
There are $512\times614$ pixels in it. Each pixel is recorded at
$224$ channels ranging from $0.38$ to $2.5\mu m$. The spectral
resolution is up to $9.46nm$. Since this hyperspectral image is too
complex to get the ground truth, we consider a subimage of $100\times100$
pixels. The first pixel starts from the $\left(105,269\right)$-th
pixel in the original image. After removing the channels $1$--$3$,
$108$--$112$, $154$--$166$ and $220$--$224$ (due to dense water
vapor and atmospheric effects), we remain $198$ channels (this is
a common preprocess for HU analyses). There are four \emph{endmembers}
latent in this data: `\#1 Tree', `\#2 Soil', `\#3 Water' and `\#4
Road', as shown in Fig.~\ref{fig:realHyperImages_japser}. 

\textbf{Urban} is one of the most widely used hyperspectral data used
in the HU area~\cite{LiuXueSong_2011_TGRS_ConstrainedNMF,Qian_11_TGRS_NMF+l1/2,Jia_09_TGRS_ConstainedNMF}.
There are $307\times307$ pixels in it, each of which corresponds
to a $2\times2\, m^{2}$ area. In this image, there are $210$ wavelengths
ranging from $0.4$ to $2.5\mu m$, resulting in a spectral resolution
of $10\, nm$. After the channels $1$--$4$, $76$, $87$, $101$--$111$,
$136$--$153$ and $198$--$210$ are removed (due to dense water
vapor and atmospheric effects), we remain $162$ channels. There are
four \emph{endmembers}: `\#1 Asphalt', `\#2 Grass', `\#3 Tree' and
`\#4 Roof' as shown in Fig.~\ref{fig:realHyperImages_urban}. 

\textbf{Cuprite} is the most benchmark dataset for the HU research~\cite{XLu_2013_TGRS_ManifoldSparseNMF,LiuXueSong_2011_TGRS_ConstrainedNMF,Qian_11_TGRS_NMF+l1/2,nWang_13_SelectedTopics_EDC-NMF,Jose_05_TGRS_Vca}
that covers the Cuprite in Las Vegas, NV, U.S. There are $224$ channels,
ranging from $0.37$ to $2.48\mu m$. After removing the noisy channels
($1\lyxmathsym{–}2$ and 221–224) and water absorption channels
(104–113 and 148–167)~\cite{XLu_2013_TGRS_ManifoldSparseNMF,Qian_11_TGRS_NMF+l1/2},
we remain 188 channels. In this paper, a region (cf. Fig.~\ref{fig:realHyperImages_cuprite})
of $250\times190$ pixels is considered, where there are 14 types
of minerals~\cite{Jose_05_TGRS_Vca}. Since there are minor differences
between variants of the same mineral, we reduce the number of \emph{endmembers}
to 12. Note that there are small differences in the setting\emph{
}of\emph{ endmembers} among the papers~\cite{XLu_2013_TGRS_ManifoldSparseNMF,LiuXueSong_2011_TGRS_ConstrainedNMF,Qian_11_TGRS_NMF+l1/2,nWang_13_SelectedTopics_EDC-NMF,Jose_05_TGRS_Vca}.
Thus, the results of the same method in their papers might be slightly
different from each other, as well slightly different from ours.

\subsection{Compared Algorithms}

To verify the performance, the proposed method is compared with seven
related methods. The information of all these methods are summarized
as follows:
\begin{enumerate}
\item \textbf{Our algorithm}: Data-guided Sparse regularized NMF (DgS-NMF)
is a new method proposed in this paper. 
\item Vertex Component Analysis~\cite{Jose_05_TGRS_Vca} (VCA) is a classic
geometric method. The code is available on \url{http://www.lx.it.pt/bioucas/code.htm}.
\item Nonnegative Matrix Factorization~\cite{Lee_99_Nature_NMF} (NMF)
is a benchmark statistical method. The code is obtained from \url{http://www.cs.helsinki.fi/u/phoyer/software.html}. 
\item Nonnegative sparse coding~\cite{Hoyer_02_NNSP_NMF_l1} ($\ell_{1}$-NMF)
is a classic sparse regularized NMF method. The code is available
from \url{http://www.cs.helsinki.fi/u/phoyer/software.html}. 
\item $\ell_{1/2}$ sparsity-constrained NMF~\cite{Qian_11_TGRS_NMF+l1/2}
($\ell{}_{\text{1/2}}$-NMF) is a state-of-the-art method that could
get sparser results than $\ell_{1}$-NMF. Since the code is unavailable,
we implement it.
\item Graph regularized NMF~\cite{Cai_11_PAMI_GNMF} (G-NMF) is a good
algorithm that transfer graph information latent in data to the new
representation. The code is obtained from \url{http://www.cad.zju.edu.cn/home/dengcai/Data/GNMF.html}. 
\item Local Neighborhood Weights regularized NMF~\cite{JmLiu_12_SlectedTopics_W-NMF}
(W-NMF) is a graph based NMF method. It integrates the spectral information
and spatial information when constructing the weighted graph. Since
the code is unavailable from the author, we implement it. 
\item \emph{Endmember} Dissimilarity Constrained NMF~\cite{nWang_13_SelectedTopics_EDC-NMF}
(EDC-NMF) urges the \emph{endmember} to be smooth  and different from
each other. The code is implemented by ourself.
\end{enumerate}
There is no parameter in VCA and NMF. For the other six methods, there
is mainly one parameter. In the next subsection, we will introduce
how to set the parameter for each algorithm. 
\begin{table*}[tb]
\noindent \centering{}\caption{The SADs and RMSEs, as well as their standard derivations, on the
Samson data. For each target, the results are arranged in rows, where
the \textcolor{red}{red value} corresponds to the best result, while
the \textcolor{blue}{blue value} is the second best one. (Best viewed
in color) \label{tab:samson_SAD&RMSE}}
\includegraphics[width=2.03\columnwidth]{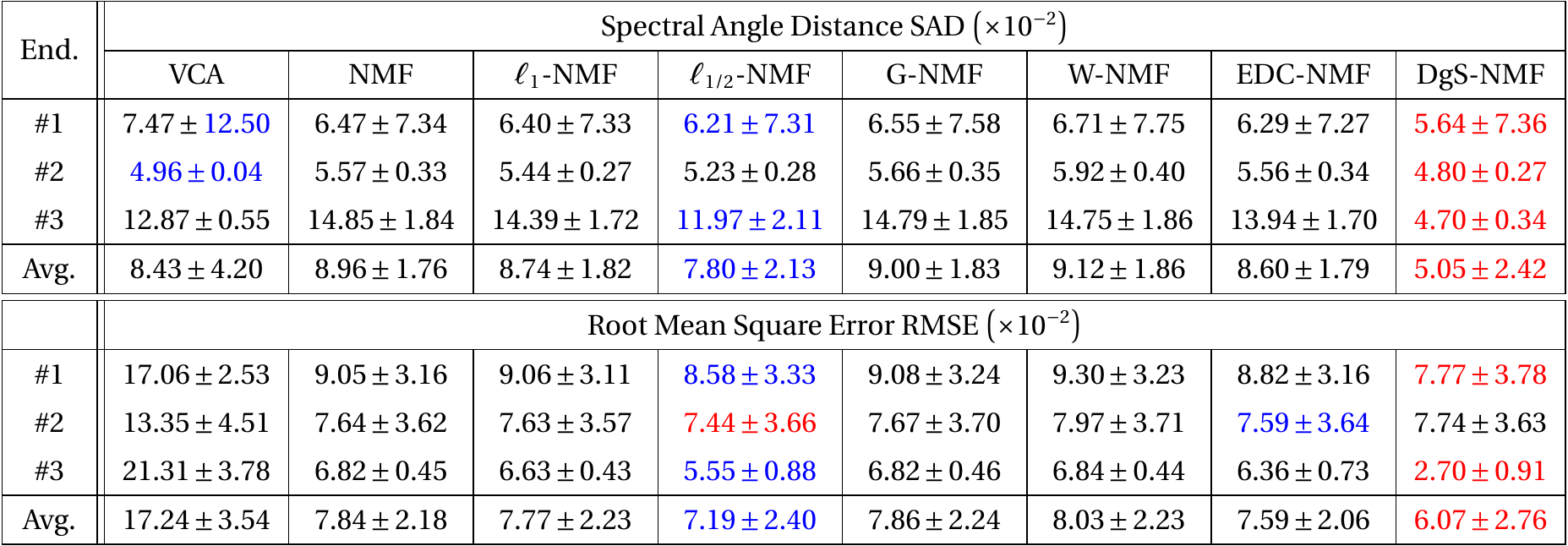}\caption{The SADs and RMSEs, as well as their standard derivations, on the
Jasper Ridge data. For each target, the results are arranged in rows,
where the \textcolor{red}{red value} corresponds to the best result,
while the \textcolor{blue}{blue value} is the second best one. (Best
viewed in color)\label{tab:jasper_SAD&RMSE}}
\includegraphics[width=2.03\columnwidth]{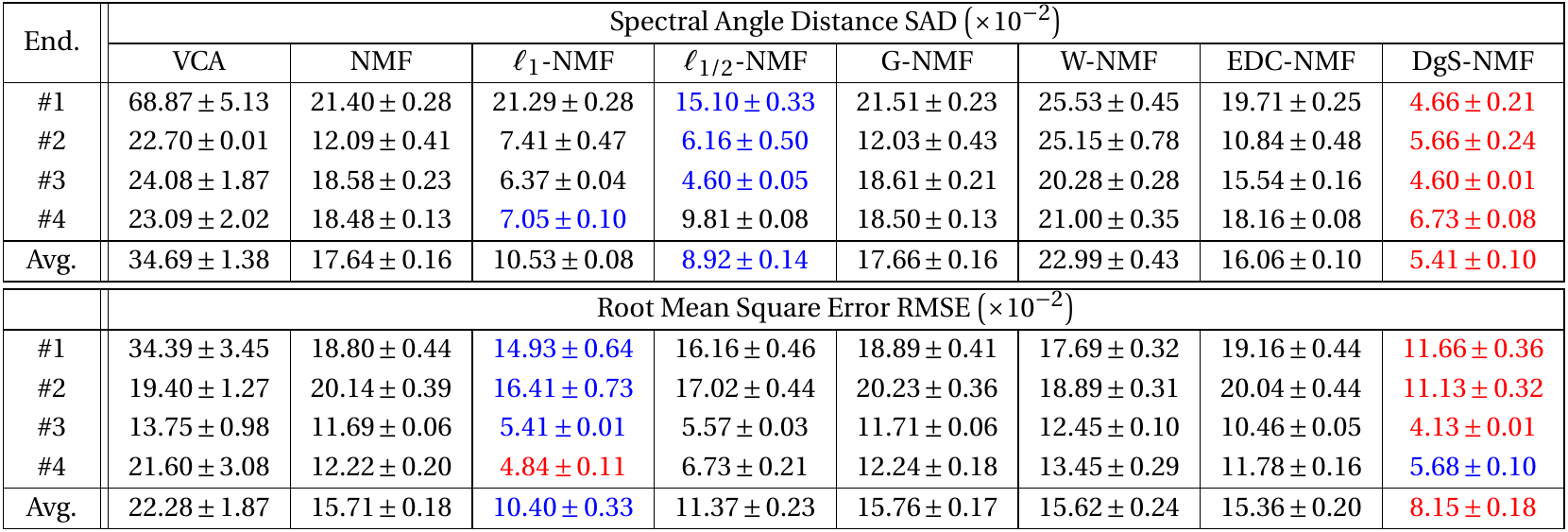}
\end{table*}

\subsection{Parameter Settings}

Similar to $\ell_{1}$-NMF and $\ell_{1/2}$-NMF, there is one essential
parameter $\lambda$ in DgS-NMF controlling the average sparsity of
the new representation. To estimate an optimal parameter, two steps
are required. First, an parameter range of $\left[\lambda_{\min},\,\lambda_{\max}\right]$
is carefully determined by trying the values at very large steps.
Second, given this parameter range, we search the best parameter by
densely searching the range of $\left[\lambda_{\min},\,\lambda_{\max}\right]$
at a number of equally spaced values. The parameter value that helps
to achieve the best result is treated as the optimal parameter setting.
For the other methods, the parameters are determined similarly. Specifically,
for our method, the optimal $\lambda$ is located in the range of
$\left[0.005,0.9\right]$ on all the datasets.

\subsection{Evaluation Metrics}

To assess the quantitative HU performance, two benchmark metrics are
introduced, i.e. the Spectral Angle Distance (SAD)~\cite{LiuXueSong_2011_TGRS_ConstrainedNMF,Jose_05_TGRS_Vca,XLu_2013_TGRS_ManifoldSparseNMF}
and the Root Mean Square Error (RMSE)~\cite{LiuXueSong_2011_TGRS_ConstrainedNMF,Qian_11_TGRS_NMF+l1/2,Kelly_2011_TGRS_SpatiAdaptiveUnmixing}.
SAD is used to evaluate the estimated \emph{endmembers}. It is defined
as 
\begin{equation}
\mbox{SAD}\left(\mtbfm,\mhbfm\right)=\arccos\left(\frac{\mtbfm^{T}\mhbfm}{\|\mtbfm\|\cdot\|\mhbfm\|}\right),\label{eq:sadMetri_experiment}
\end{equation}
where $\mhbfm$ is the estimated \emph{endmember} and $\mtbfm$ is
the corresponding ground truth. As the metric above describes the
angel distance between two vectors, a smaller SAD corresponds to a
better performance. To assess the estimated \emph{abundance}, we employ
the RMSE metric, which is given by 
\begin{equation}
\mbox{RMSE}\left(\mtbfz,\mhbfz\right)=\left(\frac{1}{N}\|\mtbfz-\mhbfz\|_{2}^{2}\right)^{1/2},\label{eq:rmseMetri_experiment}
\end{equation}
where $N$ is the number of pixels in the image, $\mhbfz$ (a row
vector in the \emph{abundance} matrix $\mhbfA$) is the estimated
\emph{abundance} map, and $\mtbfz$ is the corresponding ground truth.
In general, a smaller RMSE corresponds to a better result.
\begin{table*}[tb]
\caption{The SADs and RMSEs, as well as their standard derivations, on the
Urban data. For each target, the results are arranged in rows, where
the \textcolor{red}{red value} corresponds to the best result, while
the \textcolor{blue}{blue value} is the second best one. (Best viewed
in color) \label{tab:urban_SAD&RMSE}}

\begin{centering}
\includegraphics[width=2.03\columnwidth]{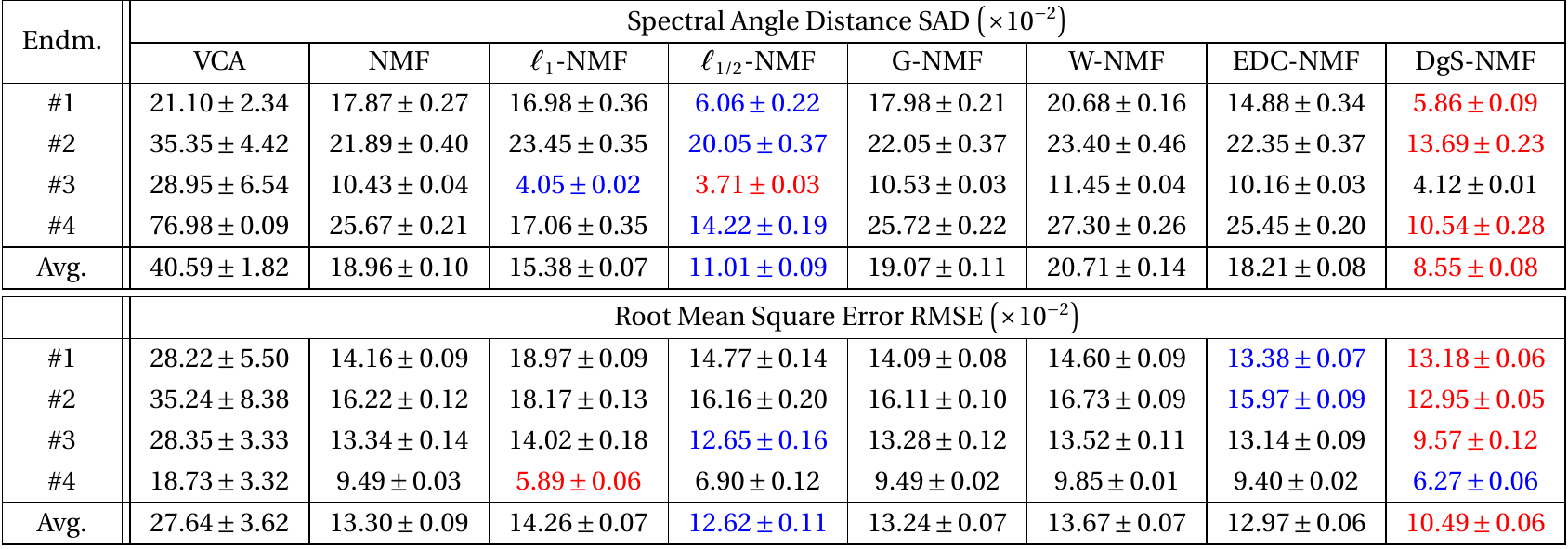}
\par\end{centering}

\caption{The SADs and their standard derivations of 6 methods on the Cuprite
dataset. There are 12 kinds of \emph{endmembers}. For each \emph{endmember},
the results are arranged in rows, where the \textcolor{red}{red value}
is the best one. (Best viewed in color) \label{tab:Cuprite_SADs}}

\centering{}\includegraphics[width=1.8\columnwidth]{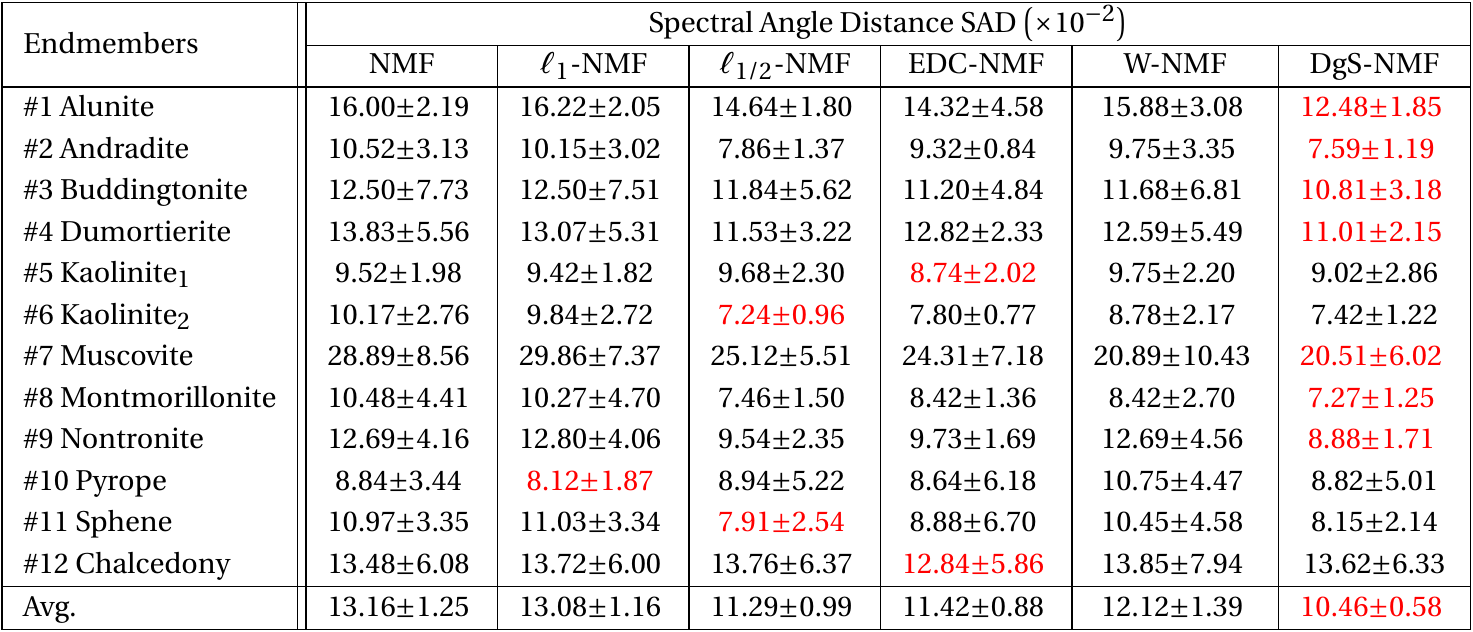}
\end{table*}

\subsection{Performance Evaluation }

To verify the performance of our method, eight experiments are carried
out. Each experiment is repeated 20 times. The mean results as well
as their standard deviations are reported. The evaluation includes
two parts: quantitative comparisons and visual comparisons.

\subsubsection{Quantitative Comparisons}

The quantitative results are summarized in Tables~\ref{tab:samson_SAD&RMSE},~\ref{tab:jasper_SAD&RMSE},~\ref{tab:urban_SAD&RMSE},~\ref{tab:Cuprite_SADs}
and plotted in Fig.~\ref{fig:AveragedPerformanceOn3DataSets}. In
Table~\ref{tab:samson_SAD&RMSE}, there are two sub-tables that show
SADs and RMSEs respectively on Samson. In the sub-table, each row
shows the performances of one \emph{endmember}, i.e. `\#1 Soil', `\#2
Tree' and `\#3 Water' in sequence. The last row shows the average
performance. In each category, the value in the red ink is the best,
while the blue value is the second best. As Table~\ref{tab:samson_SAD&RMSE}
shows, our method generally achieves the best results, and in a few
cases it achieves comparable results with the best results of other
methods. Such case is better illustrated in the $1^{\text{st}}$ subfigure
of Fig.~\ref{fig:AveragedPerformanceOn3DataSets}, where DgS-NMF
is the best method that reduces $35.3\%$ for $\overline{\text{SAD}}$
and $15.6\%$ for $\overline{\text{RMSE}}$ according to the results
of the second best method, i.e. $\ell_{1/2}$-NMF. 
\begin{figure*}[tb]
\noindent \centering{}\includegraphics[width=1.81\columnwidth]{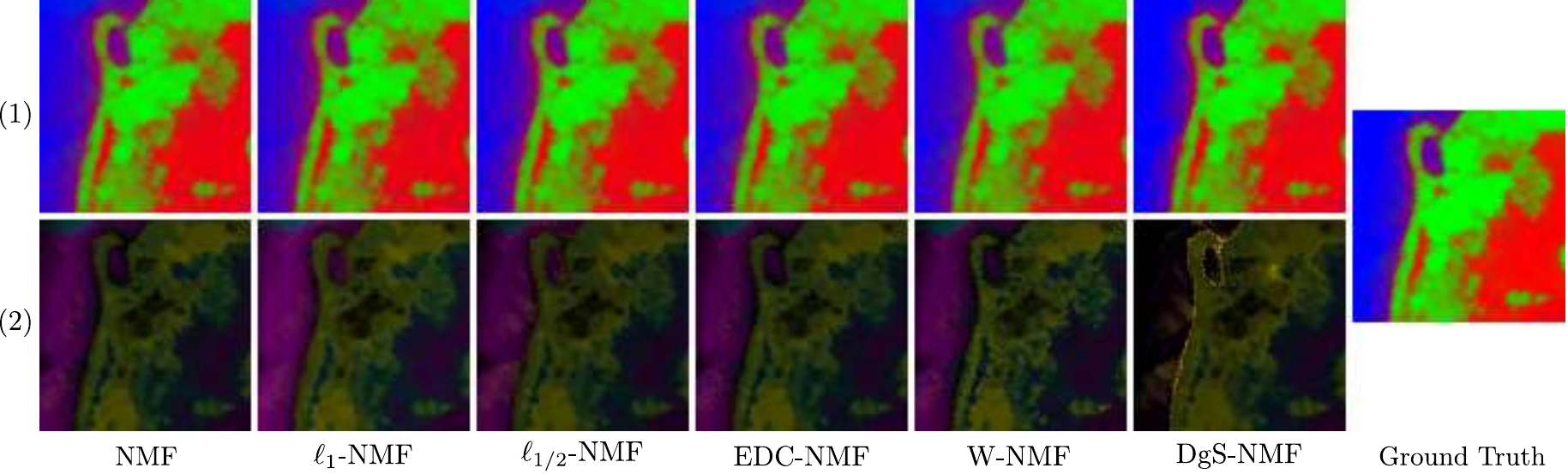}\caption{The\emph{ abundance} maps in pseudo color on the Samson data. There
are seven columns and two rows in this figure. From the  $1^{\text{st}}$
to the $6^{\mtth}$ column, each column shows the result of one algorithm.
The last column shows the ground truth. The second row shows the absolute
difference between the estimated result $\mhbfA$ and the ground truth
$\mtbfA$, i.e. $\left|\mtbfA-\mhbfA\right|\in\mtbbR_{+}^{K\times N}$.
For each subfigure, the proportions of Red, Green and Blue inks associated
with each pixel represent the \emph{abundances} of `Soil', `Tree'
and `Water' in the corresponding pixel. (Best viewed in color) \label{fig:AbundancemapsOnSamson}}
\end{figure*}
\begin{figure*}[tb]
\noindent \begin{centering}
\subfloat[\emph{Abundance} maps in pseudo color. \label{fig:AbundancemapsOnJasper-1}]{\noindent \begin{centering}
\includegraphics[width=1.81\columnwidth]{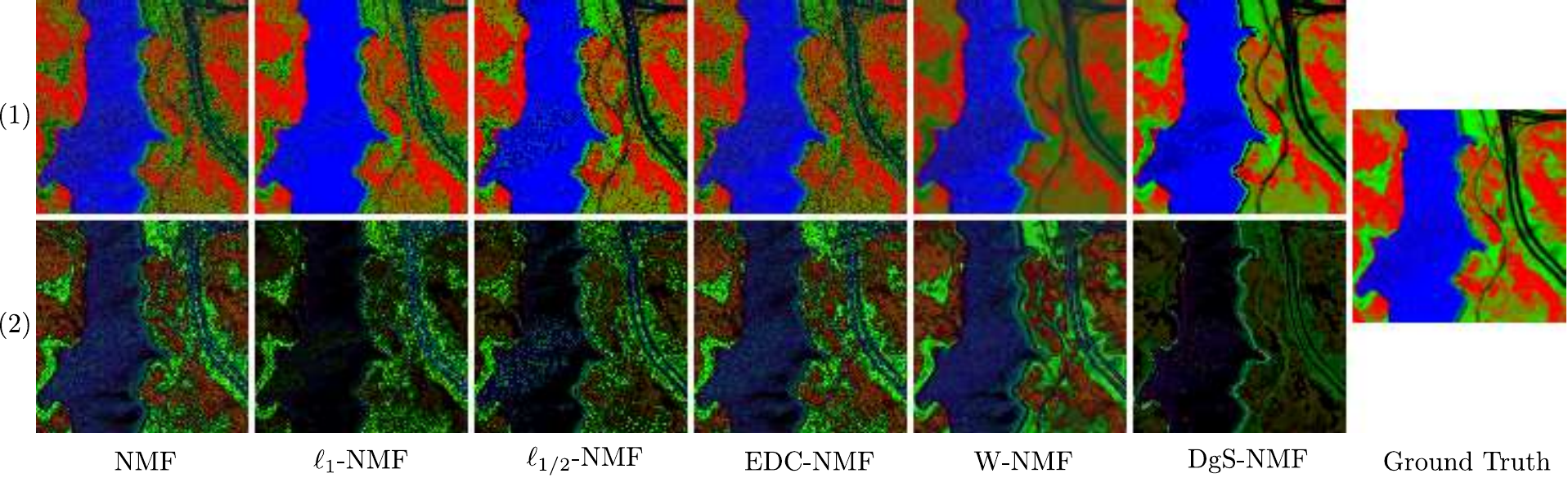}
\par\end{centering}

}
\par\end{centering}

\noindent \begin{centering}
\subfloat[\emph{Abundance} maps in gray scale. \label{fig:AbundancemapsOnJasper-2}]{\noindent \begin{centering}
\includegraphics[width=1.81\columnwidth]{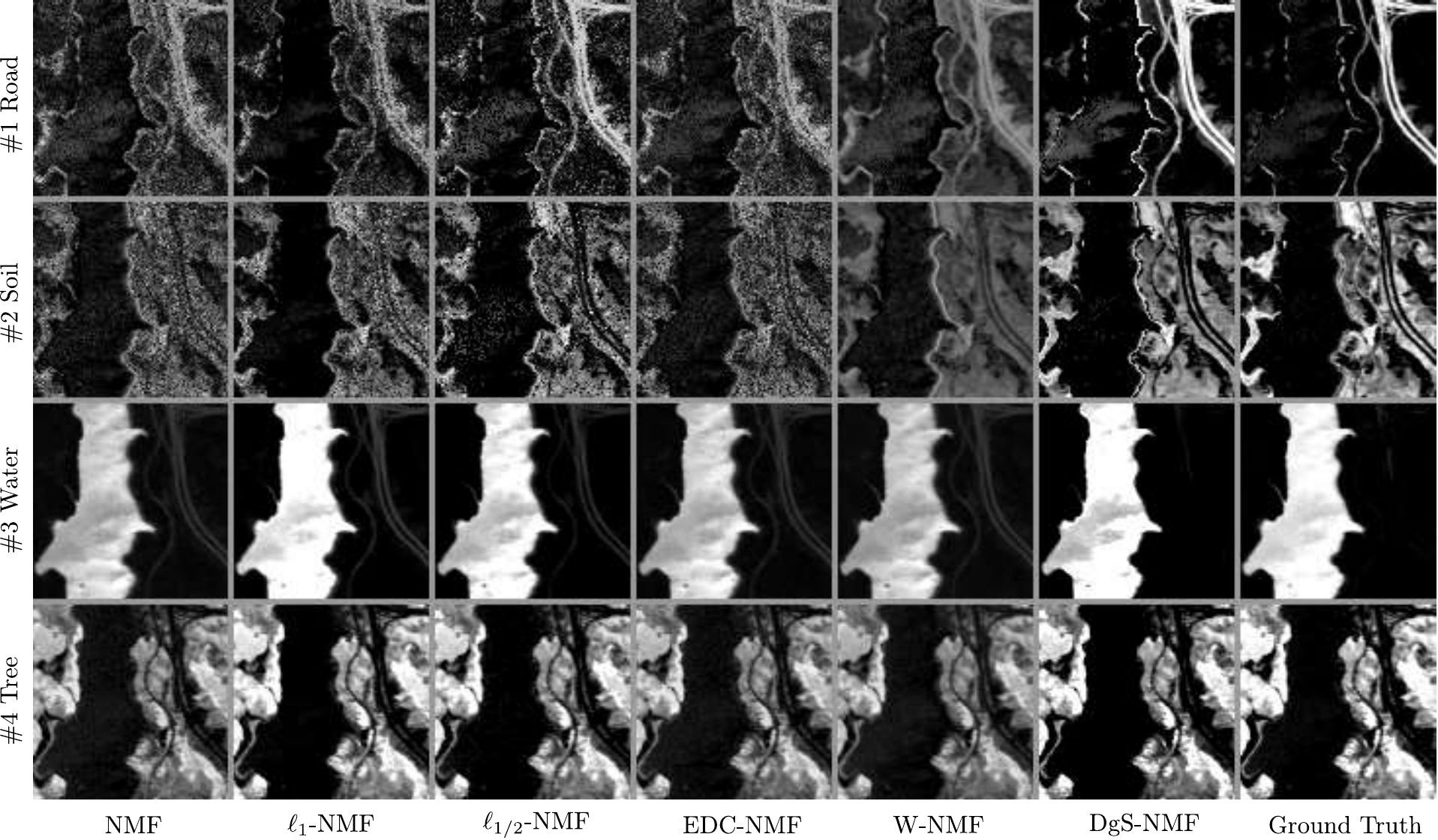}
\par\end{centering}

}
\par\end{centering}

\caption{The\emph{ abundance} maps on the Jasper Ridge data: (a) in pseudo
color and (b) in gray scale. There are two rows in (a). The second
row shows the absolute difference between the estimated result $\mhbfA$
and the ground truth $\mtbfA$, i.e. $\left|\mtbfA-\mhbfA\right|\in\mtbbR_{+}^{K\times N}$.
For each subfigure in (a), the proportions of Red, Blue, Green and
Black inks associated with each pixel represent the fractional \emph{abundances}
of `Tree', `Water', `Soil' and `Road' in the corresponding pixel.
There are four rows and seven columns in (b). Each row shows the \emph{abundance}
maps of one target. From the $1^{\text{st}}$ to the $6^{\mtth}$
column, each column illustrates the results of one algorithm. The
last column shows the ground truth. (Best viewed in color) \label{fig:AbundancemapsOnJasper}}
\end{figure*}
\begin{figure*}[tb]
\noindent \begin{centering}
\subfloat[\emph{Abundance} maps in pseudo color. \label{fig:AbundancemapsOnUrban-1}]{\noindent \begin{centering}
\includegraphics[width=1.81\columnwidth]{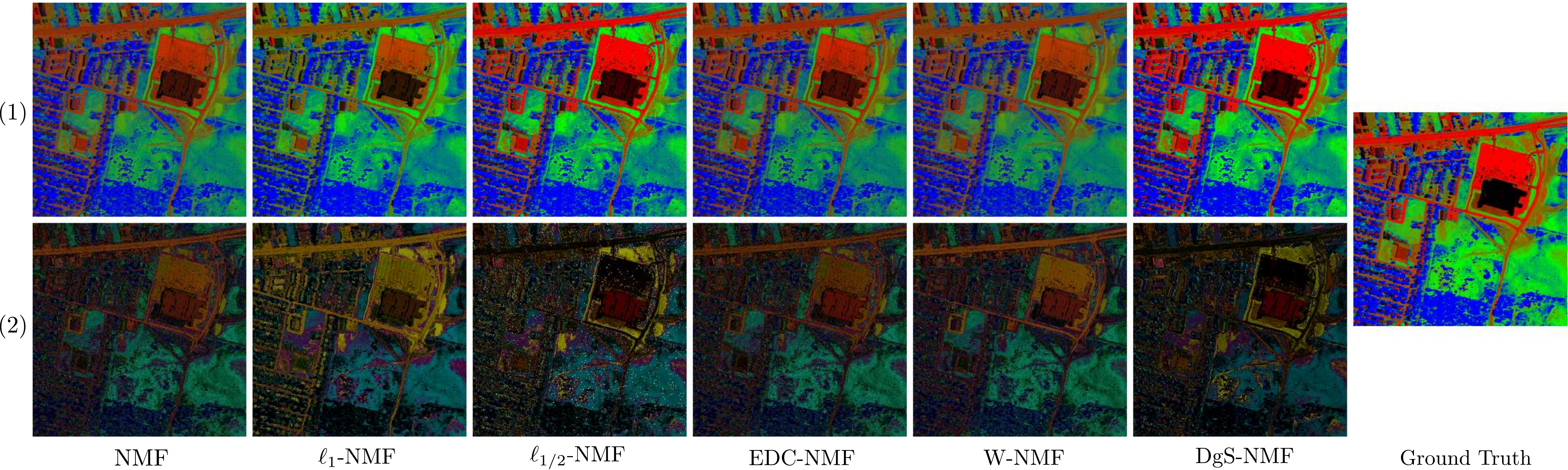}
\par\end{centering}

}
\par\end{centering}

\noindent \begin{centering}
\subfloat[\emph{Abundance} maps in gray scale. \label{fig:AbundancemapsOnUrban-2}]{\noindent \begin{centering}
\includegraphics[width=1.81\columnwidth]{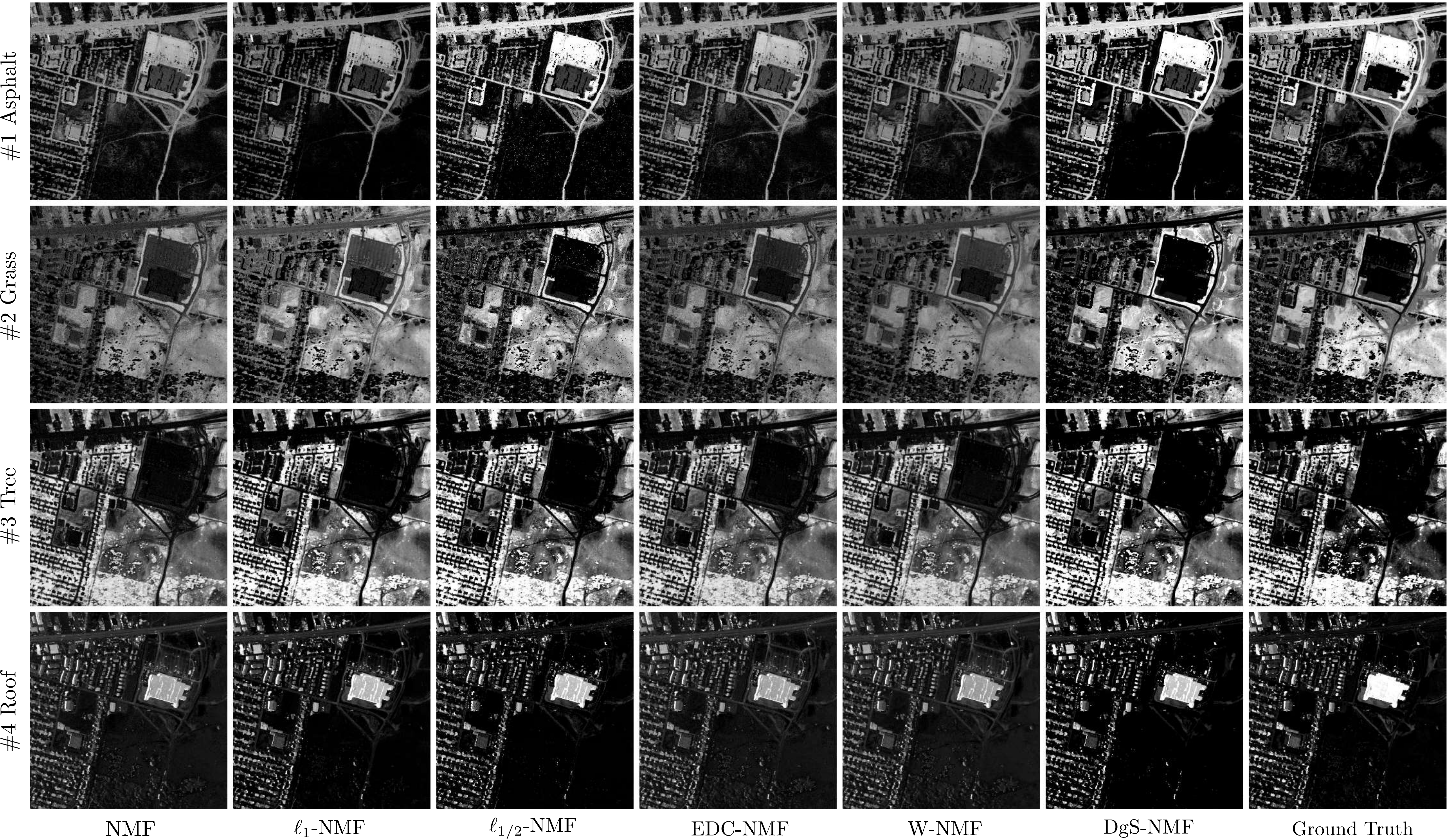}
\par\end{centering}

\noindent \centering{}}
\par\end{centering}

\caption{The\emph{ abundance} maps on the Urban data: (a) in pseudo color and
(b) in gray scale. There are two rows in (a). The second row shows
the absolute difference between the estimated result $\mhbfA$ and
the ground truth $\mtbfA$, i.e. $\left|\mtbfA-\mhbfA\right|\in\mtbbR_{+}^{K\times N}$.
For each subfigure in (a), the proportions of Red, Blue, Green and
Black inks associated with each pixel represent the fractional \emph{abundances}
of `Asphalt', `Tree', `Grass' and `Roof' in the corresponding pixel.
There are four rows and seven columns in (b). Each row shows the \emph{abundance}
maps of one target. From the $1^{\text{st}}$ to the $6^{\mtth}$
column, each column illustrates the results of one algorithm. The
last column shows the ground truths. (Best viewed in color) \label{fig:AbundancemapsOnUrban}}
\end{figure*}

Table~\ref{tab:jasper_SAD&RMSE} summaries the performances of eight
methods on Jasper Ridge. The rows show the results of four targets,
i.e. `\#1 Road', `\#2 Soil', `\#3 Water' and `\#4 Tree' respectively.
Generally, the sparsity constrained methods, i.e. $\ell_{1}$-NMF,
$\ell_{1/2}$-NMF and DgS-NMF, achieve better results than other methods.
This is since sparse constraints tend to find expressive \emph{endmembers}~\cite{Stanzli_01_CVPR_locNMF},
which might be more reliable for the HU task. The average performances
(i.e. $\overline{\text{SAD}}$ and $\overline{\text{RMSE}}$) are
illustrated in the $2^{\text{nd}}$ subfigure of Fig.~\ref{fig:AveragedPerformanceOn3DataSets}.
As we shall see, our method obtains extraordinary advantages---compared
with the second best methods, i.e. $\ell_{1/2}$-NMF and $\ell_{1}$-NMF,
DgS-NMF reduces $39.3\%$ and $21.6\%$ respectively for $\overline{\text{SAD}}$
and $\overline{\text{RMSE}}$. 

The results on Urban are illustrated in Table~\ref{tab:urban_SAD&RMSE},
where the rows contain results of `\#1 Asphalt', `\#2 Grass', `\#3
Tree' and `\#4 Roof' respectively. It can be seen that apart from
our method, $\ell_{1/2}$-NMF, $\ell_{1}$-NMF and EDC-NMF generally
achieve better results than the others. However, in general, DgS-NMF
obtains the best performance. In Fig.~\ref{fig:AveragedPerformanceOn3DataSets},
the $3^{\text{rd}}$ subfigure shows the average performances. Compared
with the second best methods, i.e. $\ell_{1/2}$-NMF, our method reduces
$22.3\%$ and $16.9\%$ for $\overline{\text{SAD}}$ and $\overline{\text{RMSE}}$
respectively. 

For the former three datasets, the number of \emph{endmembers} is
small, i.e. $K$ is small. To verify the performance of our method
on a dataset with large $K$, we carry out an experiment on the Cuprite
dataset. It is worth mentioning that the Cuprite is the most important
benchmark dataset for the HU research\ \cite{Qian_11_TGRS_NMF+l1/2,nWang_13_SelectedTopics_EDC-NMF}.
As the ground truth for the \emph{abundance }is unavailable, only
the results of \emph{endmembers} are reported in Table~\ref{tab:Cuprite_SADs}.
As we shall see, our method generally obtains the best performance.
Besides, the sparsity constrained methods, i.e. $\ell_{1/2}$-NMF
and $\ell_{1}$-NMF, usually achieve relatively good results.

\subsubsection{Visual Comparisons }

In order to give an intuitive HU comparison, we illustrate the \emph{abundance}
maps in two ways: in pseudo color and in gray scale. Fig.~\ref{fig:AbundancemapsOnJasper-1}
illustrates an example of the pseudo color manner, where there are
mainly four color inks. Through these colors, we could represent the
fractional \emph{abundances} $A_{kn}$ associated with pixel $\mtbfy_{n}$
by plotting the corresponding pixel using the proportions of red,
blue, green and black inks given by $A_{kn}$ for $k\!=\!1,2,3,4$,
respectively. So, for instance, a pixel for which $A_{2n}\!=\!1$
will be colored blue, whereas one for which $A_{1n}\!=\! A_{2n}\!=\!0.5$
will be colored with equal proportions of red and blue inks and so
will appear purple. Figs.~\ref{fig:AbundancemapsOnSamson},~\ref{fig:AbundancemapsOnJasper-1}
and~\ref{fig:AbundancemapsOnUrban-1} are obtained in this way. 

For the Samson dataset, because of the high quality of all the estimated
\emph{abundance}s, the \emph{abundance} maps in gray scale might be
very similar. For this reason, we only illustrate the pseudo color
version. The results are illustrated in Fig.~\ref{fig:AbundancemapsOnSamson}.
The top row shows the \emph{abundance} maps in pseudo color, and the
bottom row shows the absolute difference between the estimated results
$\mhbfA$ and the ground truth $\mtbfA$, i.e. $\left|\mtbfA-\mhbfA\right|\in\mtbbR_{+}^{K\times N}$.
As Fig.~\ref{fig:AbundancemapsOnSamson} shows, in general, the DgS-NMF
method achieves the minimal difference according to the ground truth.

For the Jasper Ridge data, the \emph{abundance} maps in pseudo color
and in gray scale are both provided in Fig.~\ref{fig:AbundancemapsOnJasper}.
There are four targets, i.e. `\#1 Tree', `\#2 Soil', `\#3 Water' and
`\#4 Road' respectively, the fractional \emph{abundances} of which
are illustrated by the proportions of red, blue, green and black inks
associated with each pixel, as shown in Fig.~\ref{fig:AbundancemapsOnJasper}a.
As can be seen, the sparse constraint methods, i.e. $\ell_{1}$-NMF,
$\ell_{1/2}$-NMF and DgS-NMF, get better results than the other methods.
Specifically, DgS-NMF achieves extraordinary results---the absolute
difference map in the $\left(2,6\right)$-th subfigure is the minimal
one. 

In Fig.~\ref{fig:AbundancemapsOnUrban}, the \emph{abundance} maps
in pseudo color and in gray scale are shown for the Urban data. The
four targets are as follows: `\#1 Asphalt', `\#2 Grass', `\#3 Tree'
and `\#4 Roof'. The \emph{abundances} of these targets are equal to
the proportions of red, green, blue and black inks at each pixel.
Similar to the results in Figs.~\ref{fig:AbundancemapsOnSamson}
and~\ref{fig:AbundancemapsOnJasper}, our method achieves the best
result in terms of the absolute difference map as shown in the $6^{\mtth}$
subfigure in the second row in Fig.~\ref{fig:AbundancemapsOnUrban-1}.

\subsection{Influences of Varying Parameters}

To test the stability of our method, the influences of parameters
are evaluated. Nine experiments have been conducted with respect to
nine varying parameters: $\lambda$\,$=$$0.2\lambda_{0}$,\,$\cdots$,$1.8\lambda_{0}$.
Here, $\lambda_{0}$ is the optimal parameter for each algorithm;
it might be different either for different algorithms or on different
datasets. To reduce the randomness, each experiment is repeated ten
times and the mean results are reported. 
\begin{figure*}[tb]
\begin{centering}
\includegraphics[width=2.05\columnwidth]{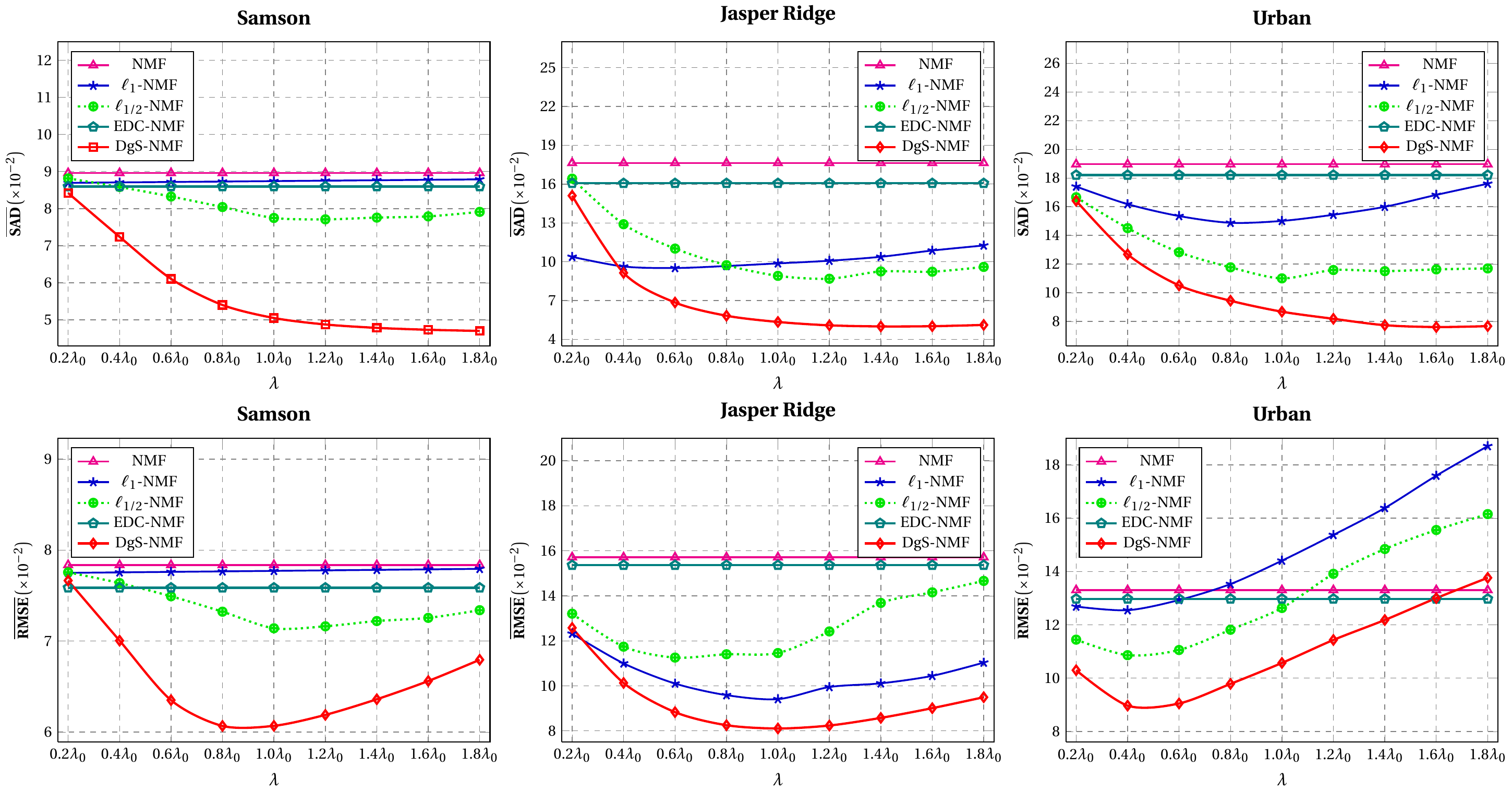}
\par\end{centering}

\centering{}\caption{Performance vs. parameter $\lambda$. There are two rows and three
columns. The top row shows $\overline{\text{SAD}}$s and the bottom
row shows $\overline{\text{RMSE}}$s. Each column shows the performance
on one dataset. In each subfigure, $\lambda_{0}$ on the X-axis donates
the best parameter setting for each algorithm. (Best viewed in color)
\label{fig:PerformCurves_vs_lambda}}
\end{figure*}

The quantitative performances are summarized in Fig.~\ref{fig:PerformCurves_vs_lambda},
where there are two rows and three columns. The top row shows the
average SADs, while the bottom row displays the average RMSEs. Each
column shows the results on one dataset. As can be seen, the curves
of NMF and EDC-NMF are plain. For the former method, there is no parameter
in it. For the latter one, we fix $\lambda$ at the optimal parameter
$\lambda_{0}$. This is because the parameter in EDC-NMF can not be
set freely; too big parameter value would lead to failure updating.
In general, the sparse constraint methods achieve better results for
all tested parameter values. Additionally, for most cases, DgS-NMF
achieves great advantages.

\subsection{\label{sub:Convergence-Study}Convergence Study}

\begin{figure*}[tb]
\begin{centering}
\includegraphics[width=2.05\columnwidth]{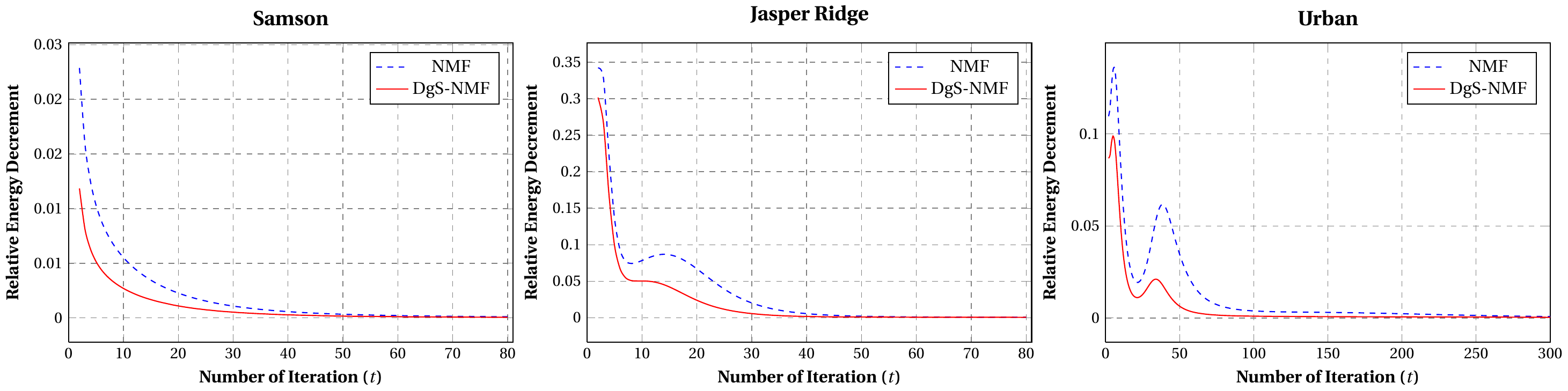}
\par\end{centering}

\caption{Relative decrement of the objective energy, i.e. $\frac{\left(\mathcal{O}_{t}-\mathcal{O}_{t+1}\right)}{\mtcalO_{t}}$,
of NMF and DgS-NMF on the three datasets: Samson, Jasper Ridge and
Urban. \label{fig:ConvergenceCurves}}
\end{figure*}

In Section~\ref{sub:ConvergenceProof_DgS-nmf}, it has been proven
that the objective~\eqref{eq:objectiveFunction_DgS-nmf} could converge
to a minimum by using the updating rules~\eqref{eq:updataDgS_M}
and~\eqref{eq:updateDgS_A}. To verify this conclusion, we study
the empirical convergence property of DgS-NMF by comparing its convergence
curves with that of NMF (a benchmark method). As shown in Fig.~\ref{fig:ConvergenceCurves},
there are three subfigures, each of which shows the results on one
dataset. In each subfigure, the X-axis shows the number of iteration
$t$, and the Y-axis illustrates the relative decrement of the objective
energy, i.e. $\frac{\left(\mathcal{O}_{t}-\mathcal{O}_{t+1}\right)}{\mtcalO_{t}}$,
of NMF and DgS-NMF. All values in Fig.~\ref{fig:ConvergenceCurves}
are nonnegative, indicating that the objective energy of both methods
decrease at each iteration. Besides, DgS-NMF converges to a local
minimum with comparable iteration steps as NMF. In this way, we've
proven Theorem~\ref{thm:nonincreasing_theorem} via empirical results.

\subsection{Influences of DgMaps }

This section gives two kinds of evaluations: 1) to evaluate the estimated
DgMaps, 2) to evaluate the contribution of the fine tuning step proposed
in Section~\ref{sub:FineTuned-DgMap}. Both visual and quantitative
comparisons have been introduced. 

Obviously, the mixed level of each pixel is closely related to the
sparse level of the corresponding \emph{abundance} vector. It is reasonable
to assess an estimated DgMap by comparing with the corresponding sparse
map of \emph{abundances} from the ground truth. Specifically, given
\emph{abundance} vectors $\left\{ \mtbfa_{n}\right\} _{n=1}^{N}\!\in\!\mtbbR_{+}^{K}$,
the $n^{\mtth}$ value in the sparse map is obtained by measuring
the sparsity~\cite{Hoyer_04_JML_NMFsparse,Qian_11_TGRS_NMF+l1/2}
of $\mtbfa_{k}$:
\begin{equation}
S_{n}=\frac{\sqrt{K}-\left\Vert \mtbfa_{n}\right\Vert _{1}/\left\Vert \mtbfa_{n}\right\Vert _{2}}{\sqrt{K}-1},\quad\forall n\in\left\{ 1,2,\cdots N\right\} ,\label{eq:sparseMap_fromGroundTruth}
\end{equation}
where $K$ is the number of elements in the \emph{abundance} vector. 

\begin{figure}[tb]
\centering{}\includegraphics[width=0.98\columnwidth]{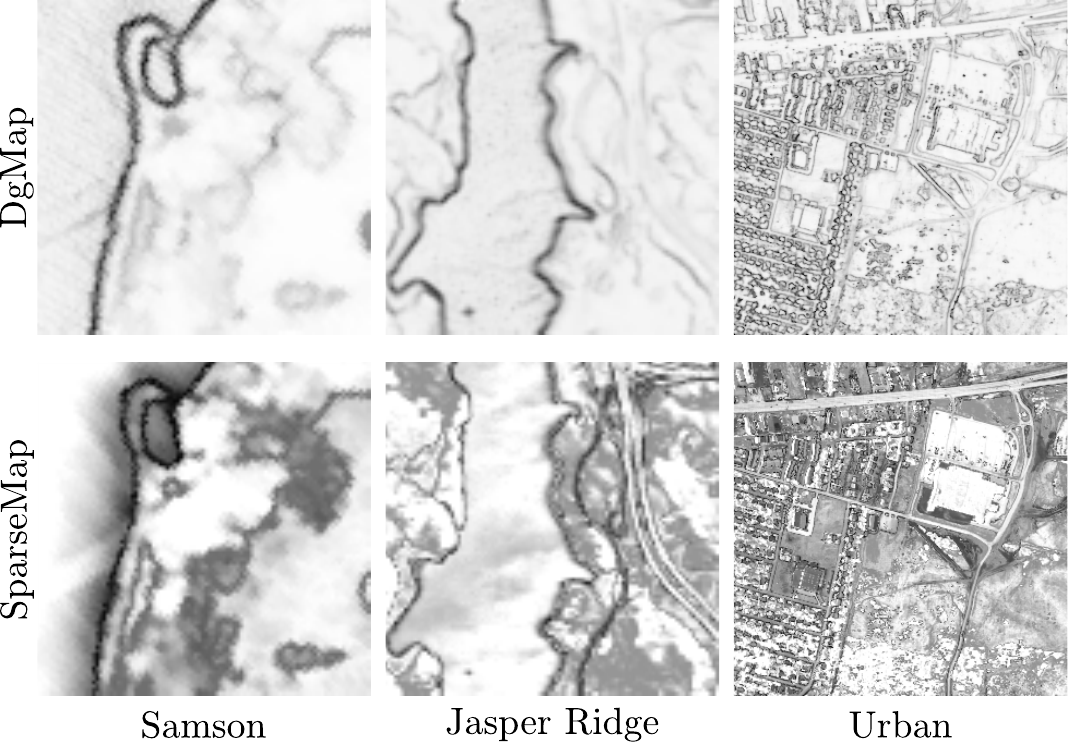}\caption{The comparison of DgMaps vs. Sparse Maps on the three datasets. DgMaps
are obtained by Section~\ref{sub:FineTuned-DgMap}; the Sparse maps
are achieved by measuring the sparsity of the \emph{abundances} from
the ground truth by~\eqref{eq:sparseMap_fromGroundTruth}. \label{fig:Comparision_DgMaps_SparseMaps}}
\end{figure}

The visual comparisons of the fine tuned DgMap and the sparse map
from ground truths are illustrated in Fig.~\ref{fig:Comparision_DgMaps_SparseMaps}.
As we shall see, the estimated DgMap is generally good. It achieves
very good results in the sudden change areas, while in the smooth
areas our method fails to capture the mixed information. 

\begin{figure}[tb]
\centering{}\includegraphics[width=0.98\columnwidth]{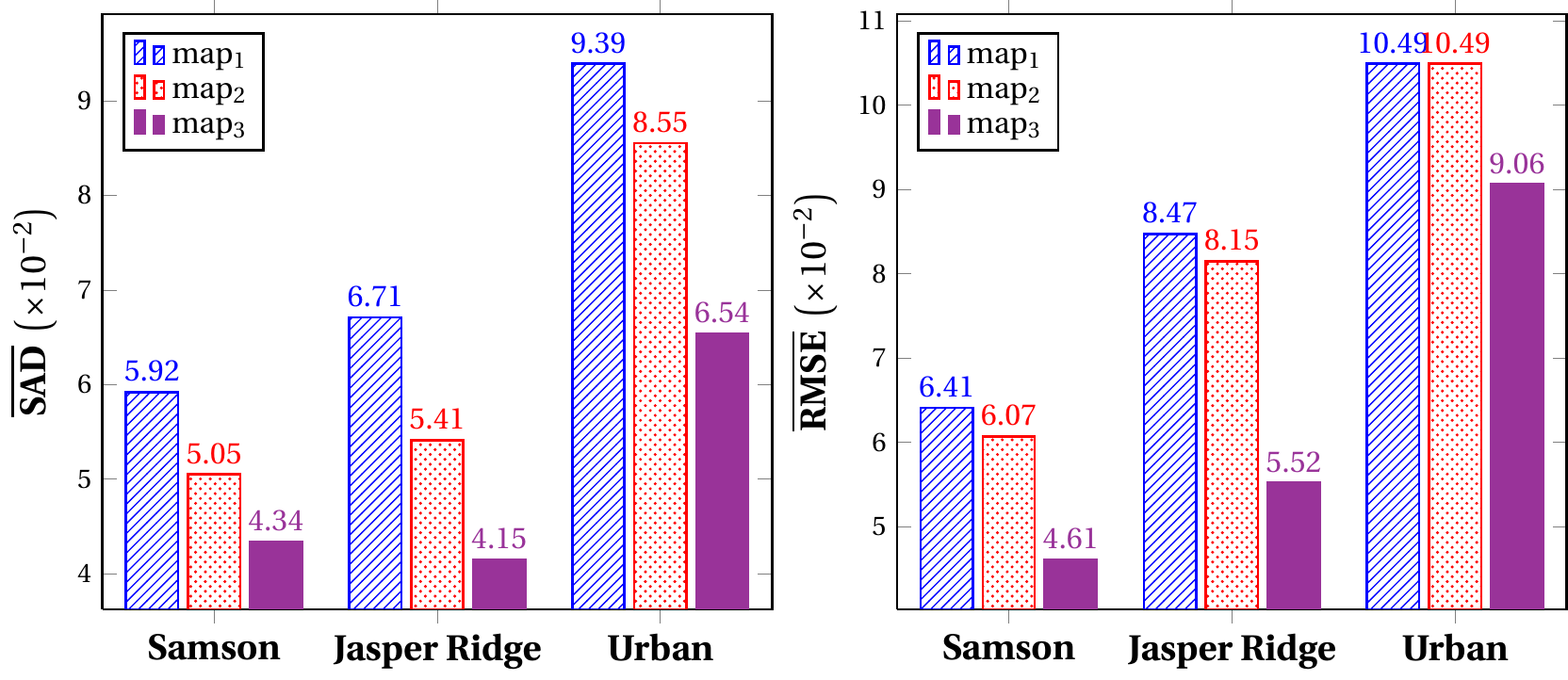}\caption{The comparison of average performances (i.e. $\overline{\text{SAD}}$
and $\overline{\text{RMSE}}$) of DgS-NMF vs. three maps, on the three
datasets. ``{\small{map$_{1}$}}'' indicates the initial DgMap; ``{\small{map$_{2}$}}''
means the fine-tuned DgMap; ``{\small{map$_{3}$}}'' denotes the sparse
map from ground truths, defined by the sparse metric\ \eqref{eq:sparseMap_fromGroundTruth}.
\label{fig:avgPerfom_2_DgMaps}}
\end{figure}

To study the quantitative evaluations, the HU performances%
\footnote{Since the standard variation of each method is similar, only the average
HU performances are provided.%
} are summarized in Table~\ref{tab:Perfom_2_DgMaps} and visualized
in Fig.~\ref{fig:avgPerfom_2_DgMaps}. There are three kinds of results
of DgS-NMF with respect to three maps: 1) ``map$_{1}$'' is the initial
DgMap; 2) ``map$_{2}$'' denotes the fine tuned DgMap; and 3) ``map$_{3}$''
means the sparse map from ground truths. As we shall see, in most
cases, the results of ``map$_{3}$'' are the best, and the results
of ``map$_{2}$'' are the second best. Such observations are better
illustrated in Fig.~\ref{fig:avgPerfom_2_DgMaps}. These observations
above imply that: 
\begin{itemize}
\item the results of the proposed data guided sparse model (DgS-NMF) is
quite promising. One can expect an even better result with a better
estimation of DgMap.
\item although the initial DgMap helps DgS-NMF to achieve good HU results,
the fine tuning process could further improve the HU performances
very much.
\end{itemize}
There are mainly two contributions of this paper. First, we propose
a data-guided sparsity model for the HU task. We have verified its
effectiveness by a heuristic DgMap estimation method. If we can obtain
a more accurate DgMap, the result can be further improved. Second,
our work introduces a new and open problem for the hyperspetral image:
how to effectively estimate a DgMap from a hyperspetral image cube?
This problem has never been considered in this area. Owing to the
encouraging result obtained by introducing the data guided sparsity,
we would like to do some further research to make it sound. The learning
based methods might be exploited to estimate better DgMaps. Besides,
the accelerating techniques used in~\cite{cbLi_2012_Tip_CompressSensing&Unmixing}
will be considered as well.
 
\begin{table*}[tb]
\centering{}{\small{\caption{The comparison of HU performances of DgS-NMF vs. three maps, on the
three datasets. ``{\small{map$_{1}$}}'' indicates the initial DgMap;
``{\small{map$_{2}$}}'' means the fine-tuned DgMap; ``{\small{map$_{3}$}}''
denotes the sparse map from ground truths, defined by the sparse metric\ \eqref{eq:sparseMap_fromGroundTruth}.
The \textcolor{red}{red value} corresponds to the best result, while
the \textcolor{blue}{blue value} is the second best result. (Best
viewed in color) \label{tab:Perfom_2_DgMaps}}
\includegraphics[width=2.03\columnwidth]{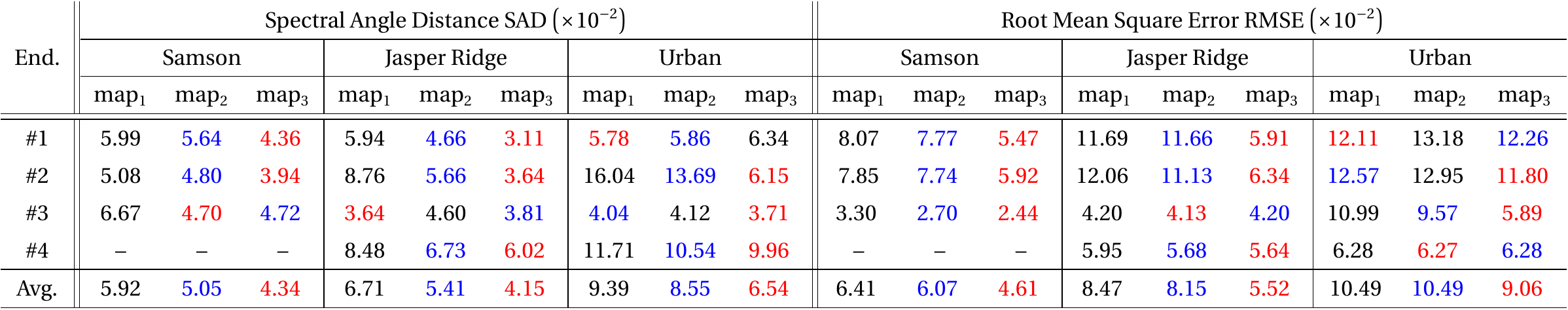}}}
\end{table*}

\section{\label{sec:Conclusions}Conclusions}

In this paper, we have provided a novel Data-guided Sparse NMF (DgS-NMF)
method by deriving a data-guided map from the original hyperspectral
image. Through this data-guided map, the sparse constraint could be
applied in an adaptive manner. Such case not only agrees with the
practical situation but also leads the \emph{endmember} toward some
spectra resembling the highly sparse regularized pixel. What is more,
experiments on the four datasets demonstrate the advantages of DgS-NMF:
1) under the optimal parameter setting, DgS-NMF achieves better results
than all the other methods in terms of both quantitative and visual
performances; 2) when the parameter varies, in most cases, our method
achieves remarkable advantages over its competitors. Besides, both
theoretic proof and empirical results verify the convergence ability
of our method.

\section*{Acknowledgements}

The authors would like to thank the editor and reviewers for their
valuable comments and suggestions. This work is supported by the projects
(Grant No. 61331018, 91338202, 61305049 and 61375024) of the National
Natural Science Foundation of China.

\bibliographystyle{IEEEtran}
\phantomsection\addcontentsline{toc}{section}{\refname}\bibliography{5F__important_doingWork_referBib_forTIP}

\end{document}